\documentclass[journal,compsoc]{IEEEtran}
\makeatletter
\def\@seccntformatinl#1{\csname the#1dis\endcsname\hskip 1em\relax}
\makeatother

\usepackage{cite, balance}
\usepackage{amsmath,amsfonts} 
\usepackage[linesnumbered]{algorithm2e}
\usepackage{notation}
\usepackage{graphicx}
\usepackage{array}
\usepackage{color}
\usepackage{graphicx}
\usepackage{float}
\usepackage{graphicx}
\usepackage{caption}
\usepackage{amsmath,amsfonts}

\def\A{{\cal A}}

\def\S{{\cal S}}
\def\T{{\cal T}}
\def\R{{\cal R}}
\def\Re{\mathbb{R}}

\DeclareMathOperator*{\argmax}{arg\,max}
\usepackage{tabularx}

\begin{document}


\title{Reinforcement Learning for Intelligent Healthcare Systems: A Comprehensive Survey } 
\author{ Alaa Awad Abdellatif$^{*}$, Naram Mhaisen$^{*}$, Zina Chkirbene$^{*}$, Amr Mohamed$^{*}$, Aiman Erbad$^{\dag}$, and Mohsen Guizani$^{*}$  \\   $^*$College of Engineering, Qatar University \\ 
$^\dag$College of Science and Engineering, Hamad Bin Khalifa University, Qatar Foundation  \\   
E-mail: \{alaa.abdellatif, aerbad\}@ieee.org,  \{naram, zina.chk, amrm, mguizani\}@qu.edu.qa   } 


\maketitle

\begin{abstract}

The rapid increase in the  percentage  of  chronic  disease  patients along with the recent pandemic  pose immediate threats on healthcare expenditure and elevate causes of death. This calls for transforming healthcare systems away from one-on-one patient treatment into intelligent health systems, to improve services, access and scalability, while reducing costs. 
Reinforcement Learning (RL) has witnessed an intrinsic breakthrough in solving a variety of complex problems for diverse applications and services.  
Thus, we conduct in this paper a comprehensive survey of the recent models and techniques of RL that have been developed/used for supporting Intelligent-healthcare (I-health) systems.  
This paper can guide the readers to deeply understand the state-of-the-art regarding the use of RL in the context of I-health. 
Specifically, we first present an overview for the I-health systems challenges, architecture, and how RL can benefit these systems. We then review the background and mathematical modeling of different RL, Deep RL (DRL), and multi-agent RL models. 
After that, we provide a deep literature review for the applications of RL in I-health systems. In particular, three main areas have been tackled, i.e.,  edge intelligence, smart core network, and dynamic treatment regimes.  
Finally, we highlight emerging challenges and outline future research directions in driving the future success of RL in I-health systems, which opens the door for exploring some interesting and unsolved problems.  


\end{abstract}
\section{Introduction \label{sec:Introduction} }

\begin{figure}[t!]
\center{\includegraphics[width=3.5in]{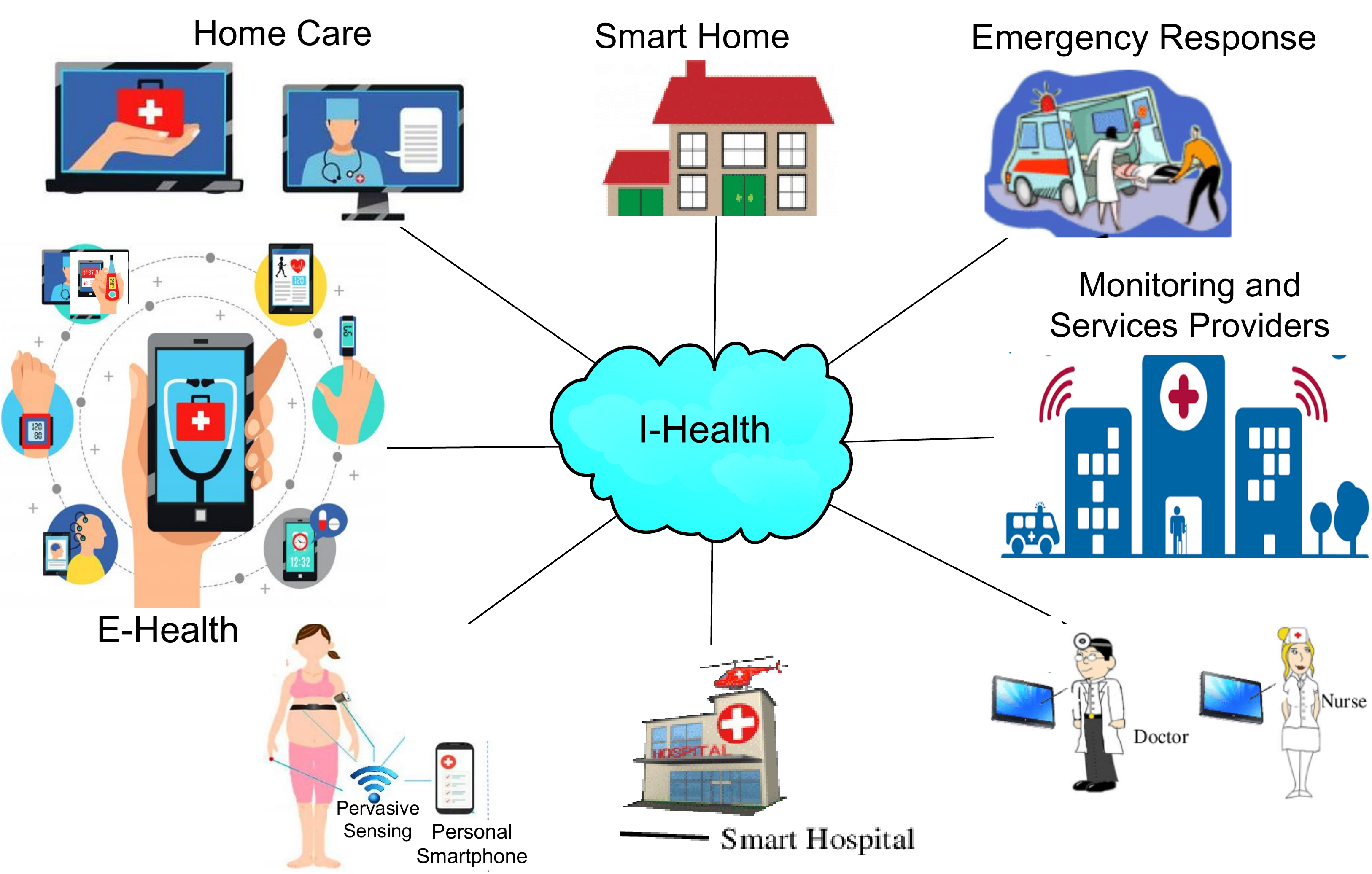}}
\caption{  I-Health system components.  }
\label{I_health2}
\end{figure}

Rapid evolution of Artificial Intelligence (AI), Internet of Mobile Things (IoMT), software defined networks (SDNs), and big data is paving the way for the emergence of Intelligent-Health (I-Health) systems. 
As healthcare is a top priority worldwide, more frameworks integrating these technologies are foreseen to be realized soon  \cite{SmarConnecHealth}. 
Leveraging ubiquitous sensing, heterogeneous 5G network, intelligent processing and control systems, I-Health systems can in real-time monitor people's daily life and provide intelligent healthcare services to both citizens and travelers without limiting their activities. I-Health can offer various applications, including remote monitoring, pandemic management, home care, and remote surgery (see Figure \ref{I_health2}).    
However, enabling high-quality healthcare services to the citizens imposes major challenges due to the rapid increase in the number of elderly  and chronic disease patients. In particular, the average age of populations around the world is rising fast, so demands for healthcare are  becoming even more significant. An aging population means an increase in the need for healthcare and treatment of chronic and age-related health issues such as:  heart disease, cancer, chronic lower respiratory diseases and diabetes. All these health issues related to a booming population segment will invariably place a high demand on the world's healthcare systems that offer a good quality of services including  the low cost, risk, and high safety of patients. 

\begin{figure*}[!h]
\center{\includegraphics[width=5.6in]{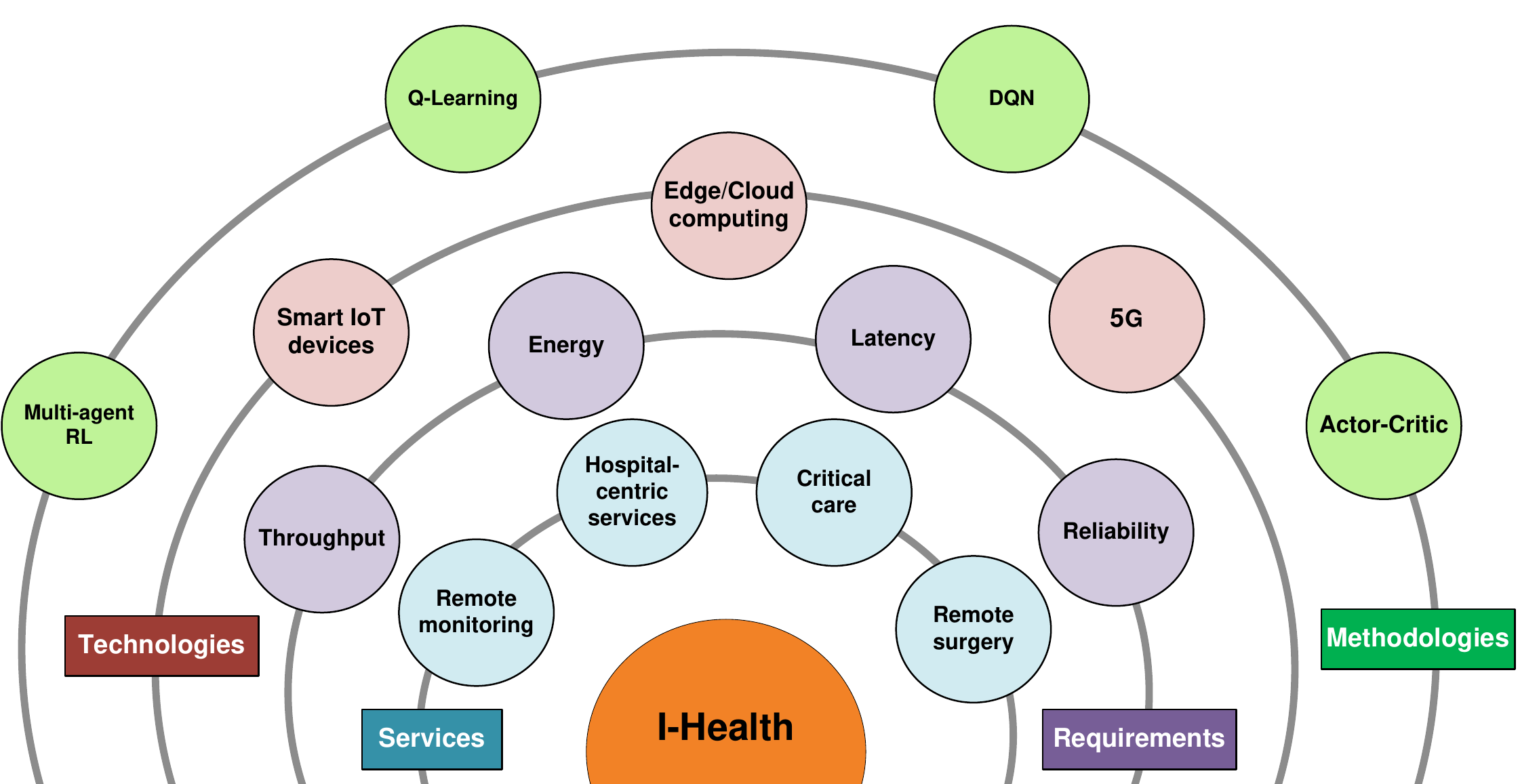}}
\caption{  I-Health system overview. The layers, from the center out, represent the services/use cases that can be supported, the requirements of the supported services, technologies that enable the implementation of the considered services, and various RL methodologies that allow the I-Health system to satisfy the different requirements.  }
\label{fig:methodology}
\end{figure*}

To address such challenges, the integration of information technology, IoMT, and modern wireless communication technologies, such as 5G, cloud computing, and edge/fog computing, provide a strong leverage to deliver a seamless, fast, and uninterrupted service to citizens \cite{9076126}. 
Typically, I-Health systems comprise a diverse number of IoMT devices that generate enormous amount of  data to intensively monitor the patients' state and automate emergency and intervention measures. These data should be processed, stored, and accessed anytime anywhere, which increases the importance of cloud computing and edge computing to reduce response time and lower bandwidth cost \cite{8382164}. However, this imposes a significant load on the system design to handle millions of devices and process such massive amounts of data. Hence, several techniques have been proposed to analyse and manage these data efficiently, such as  Reinforcement Learning (RL) techniques, which have been proposed as a promising approach for building such complex I-health systems.

In this paper, we will mainly focus on the potential applications of RL in I-Health systems. As a sub-field in AI, RL has emerged to study the optimal sequential decision-making process for an agent in non-deterministic environments. RL has obtained major theoretical and technical evolution in recent years, given its rising applicability to real-life problems, such as self-driving vehicles \cite{AutonomousDriving_drl}, robotics control, computer vision, bioinformatics \cite{Biological2018}, and natural language processing \cite{NLP2019}.  
Indeed, different RL models aim at obtaining the optimal policies leveraging prior experiences, without relying on any previous knowledge about the mathematical models of the biological systems.  
This turns RL to be more appealing than diverse control-based schemes for healthcare systems, since it is typically hard to define a precise model that simulates  the complex human body interactions to different administered treatments, due to nonlinear, varying, and delayed interaction between the treatments and human bodies.   
Figure \ref{fig:methodology} summarizes the different tiers, diverse services, technical requirements, technologies, and RL methodologies that will be discussed in this survey. We envision that supporting different types of services and applications in I-Health systems requires efficient integration between different technologies. For instance, to support remote monitoring applications, edge computing is a key component to realize ultra-low latency, while cloud computing (or smart core network) enables the optimization of resources usage and interactions across distributed nodes. At the heart of this architecture, RL can play a crucial role to optimize the I-Health system performance, decision making, and data flows throughout the overall system.


Recently,  RL techniques are gaining much interest in healthcare systems, as they exhibit fast processing capabilities with real-time predictions. 
Indeed, RL provides a class of methods for solving complicated control problems. An RL agent  interacts with the environment to learn the optimal policies that map the environment's states to actions. By knowing the current state of the environment, the RL agent can take the needed actions that define the new state while obtaining an immediate reward (such that the long-term reward over a period of time is maximized  \cite{8653879}). 
Hence, such an agent can learn from the interactions with the environment in order to obtain near-optimal actions. Conventional RL schemes utilized a lookup table to store the obtained rewards  \cite{DBLPsurv}, however, this was sufficient for  straightforward learning models with a small state space \cite{6399230}. With increasing the action and state spaces (in complex learning problems), it would be hard to obtain the optimal policy in a reasonable time leveraging such lookup tables.   
Thus, a combination of RL with deep learning has been proposed to cope up with these limitations \cite{DRL}.    
In this context, Deep Neural Network (DNN) \cite{DBLP00667} has been integrated with RL models  to generate new models, namely Deep RL (DRL), that aims at improving the conventional RL models' performance, while handling complex control problems \cite{DBLPCasas17}. These DRL models include different variants,  such as deep policy gradient RL \cite{policy_gradient},  Deep Q-Networks (DQN) \cite{DQN}, Distributed Proximal Policy Optimization (DPPO) \cite{PPO}, and Asynchronous Advantage Actor-Critic \cite{AAAC}.    

This survey reviews different types of RL and DRL techniques and their applications in I-Health systems. To the best of our knowledge, the existing surveys mainly focus on the applications of DRL in IoT systems. Also, most of the presented works are restricted to model-free single-agent DRL methods. Additionally, the concept of I-health as a future healthcare system is relatively new and not adequately tackled in the existing literature.  
Thus, this survey will focus on other less explored RL techniques, in addition to the model-free DRL, and their applications in I-health systems, while highlighting I-health system's challenges, architecture, and future research directions.  
The list of main abbreviations used in this paper is given in Table \ref{tab:abbreviations}.  
\begin{table}
    \caption{List of abbreviations used throughout the paper. }
    \label{tab:abbreviations}
    \centering
    \begin{tabular}{c|c} 
            \hline
         \textbf{Abbreviation}      & \textbf{Description}   \\ \hline 
                 AI                 & Artificial intelligence    \\ 
                 AIoT               & Autonomous IoT   \\  
                 CPS                & Cyber-physical system   \\ 
                 DDPG               & Deep deterministic policy gradient \\
                 DDQN               & Double deep Q-network    \\  
                 DRL                & Deep reinforcement learning   \\
                 DL                 & Deep learning  \\ 
                 DQN                & Deep Q-network                         \\
                 DNN                & Deep neural network          \\
                 GAN                & Generative adversarial network   \\  
                 EEG                & Electroencephalograph           \\
                 EHR                & Electronic health records     \\ 
                 HetNet             & Heterogeneous network  \\  
                 I-Health           & Intelligent-health                \\
                 IoT                & Internet of things  \\  
                 IoMT               & Internet of medical/mobile things  \\ 
                 MARL               & Multi-agent reinforcement learning   \\ 
				 MDP                & Markov decision process \\ 
                 MEC                & Multi-access edge computing  \\
                 MCC                & Mobile cloud computing    \\ 
                 ML                 & Machine learning  \\ 
                 NAC                & Natural actor-critic               \\
                 NN                 & Neural network     \\    
                 NS                 & Network slicing   \\  
                 PEN                & Patient edge node    \\ 
                 PLS                & Physical-layer security            \\
                 QoS                & Quality of service                        \\
                 QoE                & Quality of Experience     \\  
                 RAT                & Radio access technology   \\ 
                 RL                 & Reinforcement learning   \\ 
                 SDNs               & Software defined networks    \\ 
                 SL                 & Supervised learning  \\  
                 SNR                & Signal-to-noise ratio   \\  
                 SONs               & Self-organizing networks   \\  
                 TD                 & Temporal difference           \\  
                 UAV                & Unmanned Aerial Vehicle    \\ 
                 WBAN               & Wireless body area network    \\ \hline  
    \end{tabular}
\end{table}

\begin{figure*}[!h]
\center{\includegraphics[width=7.2in]{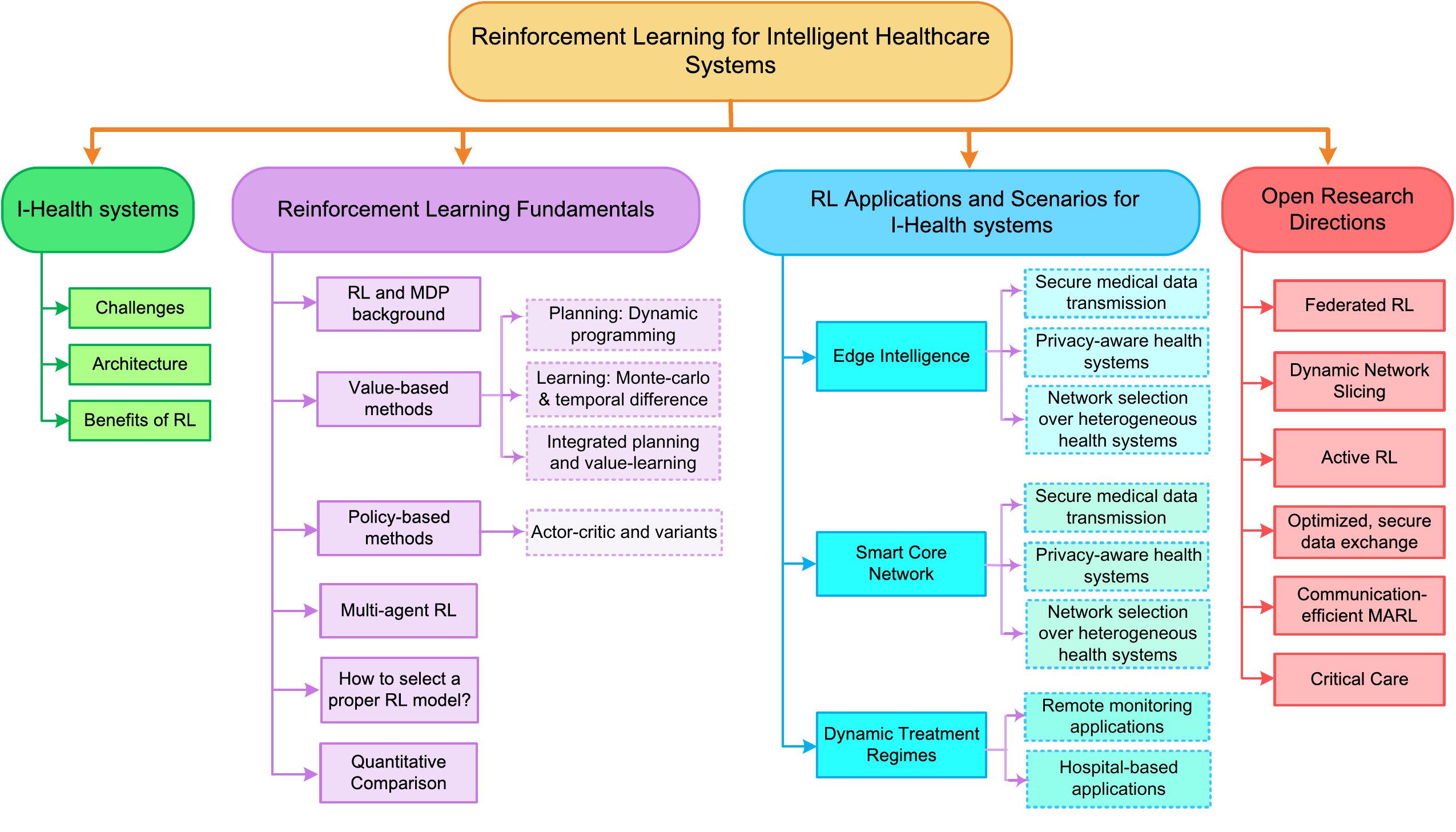}}
\caption{  A taxonomy of the reinforcement learning for I-Health systems: approaches, applications, challenges, and open research issues.    
Systems.}
\label{fig:taxonomy}
\end{figure*}

{ 

\subsection{Paper organization \label{sec:organization} } 

The rest of the paper is organized as follows.  
We first start in Section \ref{sec:Related} by discussing the related surveys that tackled RL solutions, while clearly positioning our work among others in the literature. Then, Section \ref{sec:I-Health} presents the architecture, challenges, and advantages of the considered I-Health system. 
Section \ref{sec:RL} discusses the fundamentals and background of different RL, DRL, and multi-agent RL models, while highlighting some quantitative results and tips about how to select the appropriate RL model based on the considered problem.  
Section \ref{sec:Applications} reviews the state-of-the-art approaches that adopt different RL models for optimizing the performance of healthcare systems. In particular, we categorized the presented RL applications for healthcare systems under three main areas: edge intelligence, smart core network, and dynamic treatment regimes.  
Section \ref{sec:Research}  presents our vision for the open challenges and future research directions that worth further investigation in the future. 
Finally, Section \ref{sec:Conclusion} concludes our paper. 
To sum up, Figure \ref{fig:taxonomy} illustrates how this survey is organized. 
{\color{black}

\section{Related Work \label{sec:Related} } 

\begin{table*}[htp]
	\centering
		\caption{Summary of the important papers on Reinforcement Learning}
	  \label{tab:related_work}
		\begin{tabular}{|c|c|c|c|} 
		\hline 
	\textbf{Ref.} &\textbf{Year} &\textbf{Main contribution} & \textbf{Relevance to I-health} \\
		\hline 
\cite{RL_Interpretation-Survey}	& 2020 & It reviews the main RL interpretation methods that explain the internal details & No explicit focus on I-health \\
	       &  &   of diverse RL models &  \\
	   \hline
\cite{Survey24} & 2021 & It reviews the literature on the fundamentals and applications of single and & Focusing on the enabling \\
	       &  & multi-agent DRL schemes  &  technologies for 6G networks \\
	   \hline
\cite{survey_IoT2021} & 2021 & It presents a review for the state-of-the-art DRL algorithms and their advantages and  & No explicit focus on I-health \\
	       &  & challenges in IoT applications &  \\
	   \hline
\cite{Survey11} & 2020 & It reviews the state-of-the-art DRL schemes for AIoT systems, while proposing a general  & No explicit focus on I-health \\
	       &  & model for the applications of RL/DRL in AIoT systems &  \\
	   \hline
\cite{RL_IoTSecurity2020}  & 2020 & It presents a review for the recent studies targeting securing IoT devices using & No explicit focus on I-health \\
	       &  & RL schemes &  \\
	   \hline
\cite{DRL_commNet2019} & 2019 & It reviews the applications of DRL in addressing the communications and  & No explicit focus on I-health \\
	       &  & networking problems &  \\
	   \hline
\cite{ML_6G2021} & 2021 & It presents a review for the recent machine learning based network optimization  & No explicit focus on I-health \\
	       &  & schemes that aim to optimize the end-to-end QoS and QoE  &  \\
	 \hline  
\cite{AutonomousDriving_drl} & 2021 & It presents a review for the usage of RL schemes in different autonomous  & Focuses only on autonomous  \\
	       &  & driving tasks  & driving systems \\
	 \hline 	 
	\cite{RLenergySurvey}& 2019 & It presents a brief survey about how RL-based solutions are addressing the issues & Focuses only on WBAN \\
	       &  & related to WBAN  &  \\
\hline
\cite{hh}  & 2020 & It presents an overview of RL applications for chronic diseases and critical  & Focuses only on dynamic \\
   &  & care treatments  & treatment regimes  \\
\hline
		\end{tabular}
\end{table*}
In this section, we review the main related surveys that discussed different AI schemes, including RL, in IoT and I-Health systems. Then, we highlight our contributions with respect to these related surveys.   
Table \ref{tab:related_work} recaps the important surveys that targeted RL applications, challenges, and related issues, while Table \ref{tab:related_AI} summarizes the main related surveys that discuss different related AI schemes in IoT-based systems.  

RL schemes have been widely used in different domains and applications, such as wireless resource management (e.g., power allocation, throughput maximization, and spectrum access), finance, healthcare, and automated vehicles. However, their black-box nature renders their analysis and implementation complex, especially in critical systems such as I-health systems. Thus, the authors in \cite{RL_Interpretation-Survey} review the main RL interpretation methods that are used to understand the internal design of different RL models. Moreover, several papers surveyed the potentials of DRL in different domains, such as autonomous IoT-based systems, communications and networking, 6G networks, IoT Security, and autonomous driving  \cite{Survey24, survey_IoT2021, Survey11, RL_IoTSecurity2020, Survey22,Survey23,Survey27, abdulahSurvey}. In these tutorials, the fundamentals and applications of different RL schemes have been investigated for various aforementioned systems, including rewards maximization, policy convergence, multi-agents connection, and performance optimization. However, the challenges, solutions, and open research issues of RL in I-health systems have not been reviewed before.  
For instance, the authors in \cite{Survey24} present a survey on the fundamentals and potentials of single and multi-agent DRL frameworks in future 6G networks. 
In \cite{survey_IoT2021}, a comprehensive survey for DRL algorithms in IoT applications is presented. In particular, the advantages and challenges of using DRL algorithms for a wide variety of IoT applications, such as smart grid, intelligent transportation systems, and industrial IoT applications, are discussed.  
The emerging research contributions on the applications of DRL in Autonomous IoT (AIoT) systems are summarized  in \cite{Survey11}, where a general DRL model for AIoT systems is proposed.   
Machine learning algorithms, in particular RL algorithms, have been also considered recently as one of the promising solutions for IoT security. Indeed, RL is gaining more popularity for securing IoT devices against different types of cyber-attacks. The authors in \cite{RL_IoTSecurity2020} review various types of cyber-attacks in IoT systems, while summarizing diverse RL-based security solutions presented in the literature for securing IoT devices.  

Applications of DRL in communications and networking aspects are reviewed in \cite{DRL_commNet2019, ML_6G2021}. With the rapid growth of pervasive IoT applications and Unmanned Aerial Vehicle (UAV) networks, different nodes and devices would have to make decisions locally in order to support decentralized and autonomous network functions, while maximizing the network performance. Given the high dynamics and uncertainty of diverse environments, RL has shown its efficiency in obtaining the optimal policies for different network agents (i.e., nodes) while dealing with highly dynamics networks.    
In \cite{DRL_commNet2019}, a tutorial of DRL concepts and  models is presented, while focusing on DRL schemes proposed to tackle the emerging problems in communications and networking. Such problems are trending in next generation networks, e.g., 5G and beyond, and include dynamic network association, throughput maximization, wireless caching, data/task offloading, network security, data aggregation and dynamic resource sharing, as well as connectivity preservation. 
In \cite{ML_6G2021}, the authors review the recent machine learning based network optimization
methods, implemented from the data-link layer to the application layer, to deal with the high complexity and dynamic environments in 6G networks.  
In \cite{AutonomousDriving_drl}, a review for the potentials, challenges, and advantages of employing DRL in real-world autonomous driving systems is presented.    


\begin{table*}[]
\centering
\caption{Taxonomy of the related surveys based on the discussed AI schemes.}
	  \label{tab:related_AI}
\begin{tabular}{|l|l|l|l|l|l|}
\hline
Ref              & Objective       & SL              & DL                        & RL                        & DRL                       \\ \hline
\cite{RL_Interpretation-Survey}	& \begin{tabular}[c]{@{}l@{}} It focuses on the RL interpretation methods that are used to analyze the internal design of \\ different RL models.  
\end{tabular}                 &  &                           &  \checkmark          &  \checkmark                         \\ \hline
\cite{Survey24} 	& \begin{tabular}[c]{@{}l@{}} It reviews the state-of-the-art of single and multi-agent DRL shemes and their applications  \\ in 6G networks. 
\end{tabular}                 &  &       &          &  \checkmark                         \\ \hline 
\cite{survey_IoT2021} & \begin{tabular}[c]{@{}l@{}} It review the state-of-the-art of DRL algorithms for IoT applications. 
\end{tabular}                 &  &       &          &  \checkmark                         \\ \hline 
\cite{Survey11} & \begin{tabular}[c]{@{}l@{}} It reviews the state-of-the-art of DRL schemes for AIoT systems. 
\end{tabular}                 &  &       &          &  \checkmark                         \\ \hline 
\cite{RL_IoTSecurity2020} & \begin{tabular}[c]{@{}l@{}} It presents a comprehensive survey
on the application of RL for IoT security.  
\end{tabular}                 &  &       &      \checkmark     &  \checkmark                         \\ \hline 
\cite{DRL_commNet2019} & \begin{tabular}[c]{@{}l@{}} It focuses on the applications of DRL for communications and networking problems, such as \\ network access and rate control, caching and offloading, security and connectivity preservation, \\ as well as traffic  routing and resource  scheduling.  
\end{tabular}                 &  &       &   &  \checkmark                 \\ \hline 
\cite{ML_6G2021} & \begin{tabular}[c]{@{}l@{}} It presents a survey on diverse machine learning algorithms used for supporting intelligent \\ end-to-end communication services, without focusing on I-health systems.  
\end{tabular}                 & \checkmark  &    \checkmark    &  \checkmark  &  \checkmark                 \\ \hline  
\cite{AutonomousDriving_drl} & \begin{tabular}[c]{@{}l@{}} It presents a survey on the applications of RL in autonomous driving.  
\end{tabular}                 &   &     &  \checkmark  &  \checkmark                 \\ \hline  
\cite{i6}      & \begin{tabular}[c]{@{}l@{}} This survey focuses on a general description for IoT system,\\ and wireless sensor networks. It reviews the learning \\ and big data studies related to IoT.  
\end{tabular}                 & \checkmark &                           &                           &                           \\ \hline
\cite{i10}     &  \begin{tabular}[c]{@{}l@{}} It provides an overview on smart transportation systems including\\ ML techniques and IoT applications in Intelligent\\ Transportation Systems (ITS).                                                              \end{tabular}   & \checkmark &                           &                           &                           \\ \hline
\cite{i22}     &  \begin{tabular}[c]{@{}l@{}}This survey focuses on the IoT for cloud/fog/edge \\ computing and the use of ML in MEC systems.                                                         \end{tabular}   & \checkmark &                           &                           &                           \\ \hline
\cite{dhh}     &  \begin{tabular}[c]{@{}l@{}}  It surveys the existing techniques\\ of deep learning in the fields of translational bioinformatics,\\ medical imaging, pervasive sensing, medical informatics and public health. \end{tabular}       & \checkmark & \checkmark &                           &                           \\ \hline
\cite{s51} & \begin{tabular}[c]{@{}l@{}} It surveys the existing DRL techniques  developed for cyber security and \\ various vital aspects, including DRL-based security models for cyber-physical systems, \\ autonomous intrusion detection schemes, and multi-agent DRL-based game theory simulations \\ for defense strategies against cyber attacks. \end{tabular}                                                           &                           &                           &                           &         \checkmark                  \\ \hline
\cite{i11}     &  \begin{tabular}[c]{@{}l@{}} It reviews the AI applications for stroke, specifically it focuses on three major areas,  \\ i.e.,  early detection and diagnosis, dynamic treatment, as well as the outcome \\ prediction and prognosis evaluation. \end{tabular}                                                                                &         \checkmark                   & &                           &                           \\ \hline
\cite{i14}     &  \begin{tabular}[c]{@{}l@{}} It reviews the related work for the implementation of AI into existing clinical workflows, \\  including data sharing and privacy, transparency of algorithms, data standardization,  and  \\ interoperability across multiple platforms, while considering the major concerns for patient safety.  \end{tabular}                                                                         &         \checkmark                  &                           &  &                           \\ \hline
\cite{Review_drl2019} &  \begin{tabular}[c]{@{}l@{}} It presents a general overview for the neural networks and deep neural networks challenges, \\  training algorithms, architecture, and  implementations, without focusing on I-health systems.  \end{tabular}       & \checkmark &   & \checkmark    &   \checkmark                          \\ \hline
\cite{RLenergySurvey}   & \begin{tabular}[c]{@{}l@{}} It investigates some of the RL and DRL solutions that targeted WBAN. 
\end{tabular}                 &  &      &   \checkmark    &     \checkmark     \\ \hline 
This survey      &   \begin{tabular}[c]{@{}l@{}} It presents a comprehensive survey about different RL and DRL applications  in I-health systems. 
\end{tabular}             &           &              &         \checkmark  &     \checkmark       \\ \hline  
\end{tabular}
\end{table*}

Although there are other surveys related to RL and DRL, most of them ignore healthcare applications and related characteristics and challenges. For instance, \cite{li2017deep} and \cite{8103164} review the usage of DRL algorithms for computer vision and natural language processing. The survey in \cite{chen2017machine} focuses on the applications of deep learning schemes in wireless networks, while the  survey in  \cite{duan2016benchmarking} focusing on the problems of  continuous control and large discrete action space.    
To the best of our knowledge, the only two surveys that partially address  some applications of RL in healthcare systems are \cite{RLenergySurvey, hh}.  
The survey in \cite{RLenergySurvey} discusses the potential of RL-based solutions in Wireless Body Area Network (WBAN). In particular, this survey focuses on the energy management issues in WBAN, which include inter-network interference management over the multi-agent environment, energy consumption minimization, power control for in-body sensors, trade-off between energy efficiency and transmission delay, as well as power control to mitigate jamming. Then, it discusses the main solutions that address these issues, such as dynamic power control, sensor access control, and energy harvesting. 
The survey in \cite{hh} discusses some of the RL models that are used in chronic and critical diseases treatments. 
This survey has focused on the applications of RL for dynamic treatment regimes, however, it ignores the edge intelligence, smart core network, as well as networking and communication aspects in healthcare systems.  
This motivates us to develop a survey that presents a comprehensive literature review on the architecture, requirements, and challenges of I-health systems, while focusing on the applications of RL to tackle these challenges and requirements in different system levels.

\subsection{Our contribution \label{sec:contribution} } 

To the best of our knowledge, this is the first survey that addresses the potentials of RL in all layers of healthcare systems.   
Specifically, the main contributions of this survey lie in the following aspects: 
\begin{itemize}
\item We first review the major challenges of I-health systems and propose a generic I-health system architecture that integrates diverse components of any I-health system. Then, we discuss why RL is needed to cope with the increasing demand of I-health systems by addressing these challenges.  
\item  We provide a comprehensive tutorial on single-agent and multi-agent DRL frameworks. In particular, we briefly explain the main concepts of diverse RL schemes and their fundamental building blocks, 
including the value-based and policy gradient models, while discussing the pros and cons of each category and how to select the appropriate RL model to use. 
Moreover, we empirically assess the performance of different value-based methods (i.e., DQN and actor-critic methods).  
\item We focus on the potentials and applications of different RL models in I-health systems, where the relevant studies are categorized into three main sectors, i.e. edge intelligence, smart core network, and dynamic treatment regimes. In each sector, we discuss different RL models that are used to fulfil various smart healthcare services' requirements, following the I-health system architecture. The proposed I-health architecture not only helps in building a taxonomy to recap and classify existing studies, but also provides a general framework to investigate the potentials of different AI techniques in I-health systems.   
\item Finally, we discuss several future research directions related to the deployment of efficient, scalable, and decentralized RL schemes in I-health systems in order to enhance the available healthcare services while enabling new, smart services for future healthcare systems. 
\end{itemize}
}
}
\section{I-Health Systems: Challenges, Architecture, and Benefits of RL  \label{sec:I-Health} }

In this section, we first present the main challenges for implementing I-Health systems. Then, the proposed I-Health system architecture is introduced to address these challenges. After that, we highlight the major benefits that can be obtained by incorporating RL schemes within the proposed I-Health architecture. 

\begin{figure*}[t!]
\center{\includegraphics[width=5.2in]{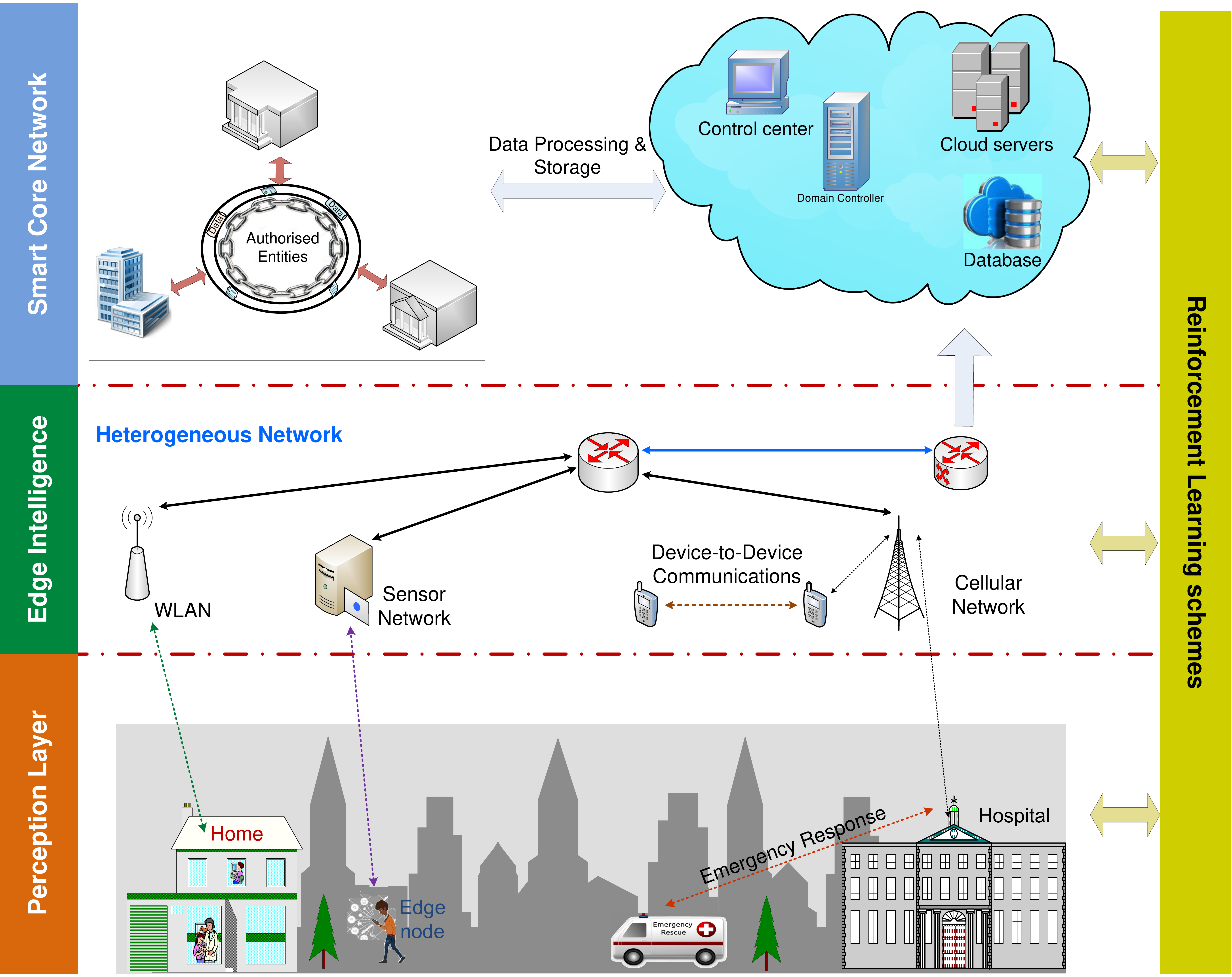}}
\caption{ The proposed I-Health system architecture.  }
\label{fig: Architecture}
\end{figure*}

\subsection{Challenges of I-Health systems}

Despite the promising evolution towards enabling remote health systems, several challenges still have to be addressed towards providing intelligent healthcare services.   
\\    
\textbf{Highly dynamic environment:}  
The evolution of e-health allows for collecting, processing, and analyzing piles of information from several devices/locations (e.g., hospitals, clinics, etc) in order to provide efficient healthcare services. The advances of edge nodes (e.g., smartphones), equipped with built-in sensors, cameras, and high-performance computing and storage resources enable each patient/user to generate massive amount of data anytime and anywhere. The patients can generate, collect, and communicate irregularly large volumes of real-time data about its own operation and surrounding environment. Thus, given the anomaly in generating and collecting the medical data in addition to the networks' dynamics, I-Health systems turn to be heterogeneous and highly dynamic environment. Moreover, such dynamics are too complex to be captured with either simple stochastic assumptions or static conventional optimization. Such highly, non-stationary environment typically calls for adopting experience-based learning.   \\     
\textbf{ Large number of potential users:}  Given the rapid increase in the number of the chronically ill and elderly people, most of the healthcare facilities are required to serve thousands of patients daily \cite{MEdge-Chain}. This number can be easily duplicated in case of
pandemic, such as the recent COVID-19 outbreak \cite{WHO, Covid_data}, which puts
a serious load on different healthcare facilities.  
Given such enormous number of patients/users in the I-Health system, providing accurate, sequential personalized clinical decisions is a very challenging task.   \\   
\textbf{ Distributed and Imbalanced data:} The heterogeneity of the I-Health system, and the distribution and size of the collected data from different patients significantly vary. The locally generated data in I-Health systems is distributed across multiple devices/nodes in the network, e.g., smartphones, sensors, hospitals, and cameras, which results in imbalanced data distribution. Most of the machine learning algorithms may suffer from biased and inaccurate prediction due to the imbalanced data. 
Thus, heterogeneous and imbalanced data along with the high-dynamic environment put major challenges in designing an efficient I-Health system that supports reliable and real-time healthcare services.  \\   
\textbf{ Limited computational and communication resources:}  The large number of patients/users participating in the I-Health system, the availability of the needed computational and
communication resources at different network levels can be challenging. The generated traffic from different locations of the I-Health system grows linearly with the number of participating users. Furthermore, the heterogeneity of the edge nodes or patients' equipment, in terms of computational capabilities and energy availability (i.e., battery level) introduces an extra constraint to optimize the performance in I-Health system.  

\subsection{I-Health system architecture}

To tackle the above challenges, we propose the I-Health architecture in Figure \ref{fig: Architecture}.   
The proposed architecture is comprised of three main layers: perception layer, edge intelligence layer, and smart core network layer. This architecture stretches from the physical layer, where data is generated/collected, to the control and management layer. It includes the following main components: \\ 
\textbf{Perception layer:} A combination of sensing and IoMT devices represents the set of data sources, which provide real-time monitoring and ubiquitous sensing for diverse E-health applications. \\  
\textbf{Edge intelligence layer:} This layer focuses on processing the acquired data from different sources (e.g., transferring it from raw sensory data into actionable insights), in addition to the association with the heterogeneous network (HetNet) infrastructure. Thus, it represents the intermediate processing, communication, and storage
stage between heterogeneous data sources and core network. In particular, an intelligent edge node can gather the medical and non-medical data from different sources, to analyze, classify, and extract information of interest, then it forwards the processed data or extracted information to the core network through HetNet \cite{comp_class}. The latter incorporates cellular networks, Wireless Local Area Networks (WLANs), Device-to-Device communications (D2D), and sensor networks. Such a heterogeneous network enables seamless switching among different types of technologies in order to optimize medical data delivery.  
Interestingly, different health-related applications (apps) can be also implemented in such intelligent edge nodes. These apps can play crucial role in long-term chronic diseases management, while assisting patients to actively involve in their treatment regime by ubiquitously interacting with their doctors anytime and anywhere.  \\   
\textbf{Smart core network:} It takes comparative advantages of powerful computing sources, i.e., from the edge to the cloud, to process, analyze, and store the collected data. Moreover, authorized entities, such as the government and big organizations, can have certain privilege and authorization at this layer to access the collected information and define the requirements or policies for decision-making and control. We highlight that the proposed architecture enables the two-way communication, i.e., sensing and control. This two-way communication enables acquiring the knowledge about the physical world, while monitoring and managing every device/component in the system to make it operate properly and smartly. Hence, leveraging the acquired data from the physical layer, different RL schemes can be implemented at each layer of the proposed architecture to facilitate various decision-making processes. 

\subsection{Why RL is needed for I-Health systems?} 

The gathered data at an I-Health system can be processed at different system levels (i.e., edge devices, fog nodes, and the cloud) from perception layer to core network layer. Thus, different RL schemes can be represented by a transversal layer, i.e., a vertical layer crossing over all layers (see Figure \ref{fig: Architecture}), where they can be implemented to efficiently process the gathered data and configure optimally the different parameters at each layer. Thus, the main motivations of applying RL in I-Health systems can be summarized as follows: 
\begin{itemize}
    \item Optimizing data processing and transmission in I-Health systems.  Leveraging RL allows for optimizing the integration of distributed data processing and communication in the resource-constrained environment, such as I-Health systems. The ability of RL to derive optimal decision-making strategies from observational data without having a model for the underlying process makes it a promising approach in optimizing I-Health systems' behaviour.
    \item Defining optimal dynamic treatment regimes. 
    One of the main targets in I-Health system is to provide effective treatment regimes that: (i) dynamically adapt to the varying patients' state, (ii) enable personalized medication through responding to diverse patients' responses to various treatment regimes, and (iii) enhance the long-term benefits of patients. RL framework perfectly fits in designing optimal dynamic treatment regimes that can be viewed as sequential decision-making problems.  
    Each agent in the RL framework aims to learn the optimal policy from the interactions with the environment in order to maximize its reward. This is precisely mapping to what happens in dynamic treatment regimes, where the set of rules followed in dynamic treatment regimes are equivalent to the policies in RL, while the treatment outcomes are equivalent to the rewards in RL.  
    Thus, RL is an appealing approach in designing adaptive clinical strategies for the treatment of diverse chronic disease \cite{vincent2014reinforcement}. 
    \item Supporting automated medical diagnosis processes. Driven by the recent advances in big data analysis and machine learning schemes, there is much interest to promote the diagnostic process toward enabling automated medical diagnosis \cite{drl_diognose2020, drl_diognose2017}. In order to provide error-resistant process in diagnosis while assisting the clinicians for an effective decision making, several studies work on formulating the diagnostic classification problem as a sequential decision-making process. This allows for leveraging the RL in solving such dynamic problem with a small amount of labeled data generated from relevant resources \cite{drl_diognose2019}.   
\end{itemize}

\section{Reinforcement Learning: Overview and Theoretical Background  \label{sec:RL} }  

\begin{figure}[t!]
\center{\includegraphics[width=3.4in]{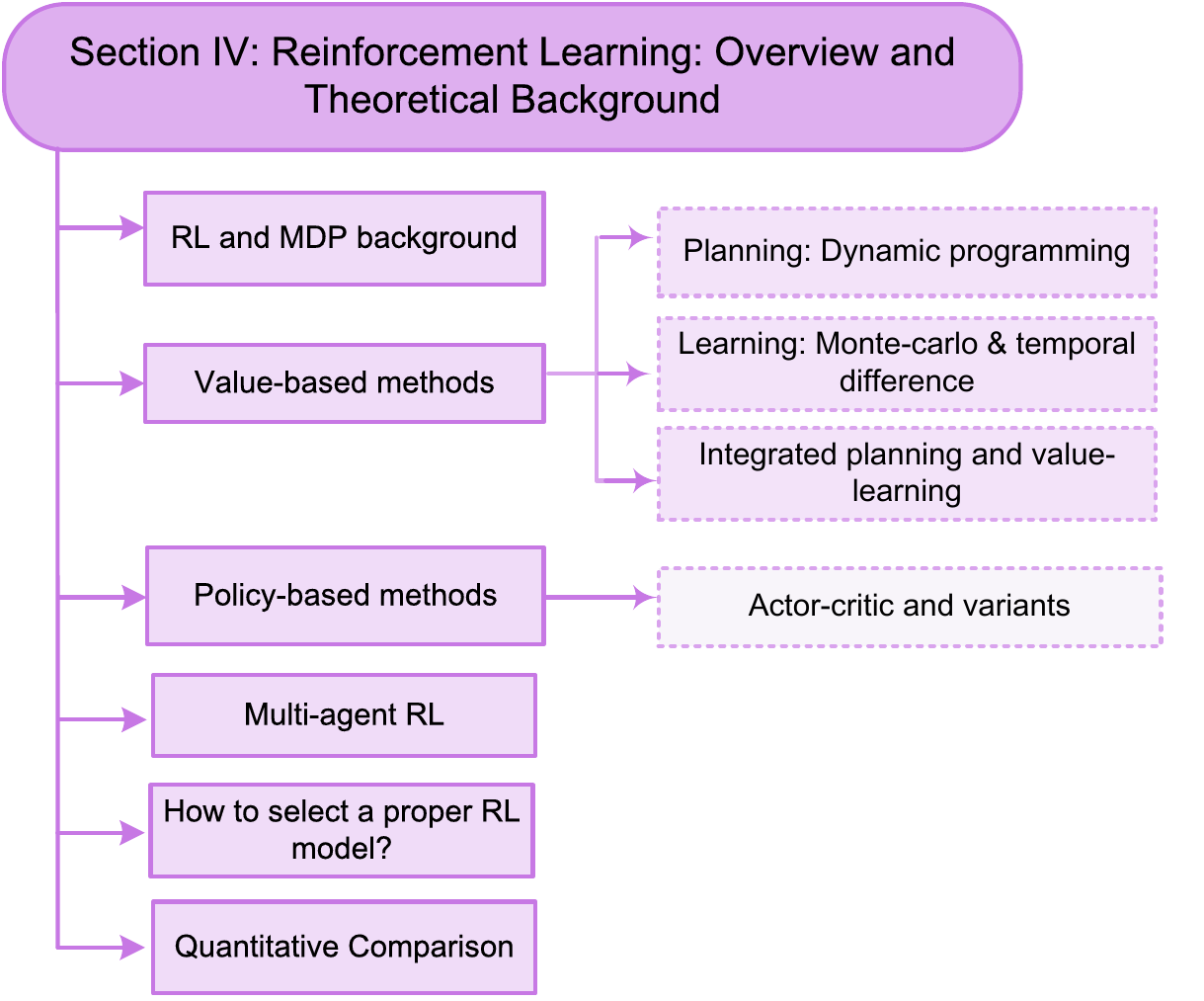}}
\caption{ Taxonomy of RL fundamentals section.  }
\label{fig: RLsection}
\end{figure}

This section presents a brief overview to the theoretical background, fundamental concepts, as well as different models and advanced techniques in RL. Subsection \ref{sec:RL1} first describes the general model of Markov Decision Processes and how RL can be used to deduced optimal polices for that model. Then, Subsection \ref{sec:RL2} discusses the main value-based methods that are used to solve the MDP problem, while Subsection \ref{sec:RL3} investigates different policy gradients-based methods that search directly in the space of policies to find the optimal policy. Subsection \ref{sec:RL4} discusses the motivation of leveraging MARL framework to model and solve diverse control problems, while highlighting the main proposed algorithms in this area. In Subsection \ref{sec:RL5}, we propose a generic way to identify how to select the appropriate RL model to use based on the targeted application's requirements. Finally, we present some quantitative comparisons in Subsection \ref{sec:RL6} in order to empirically depict the difference between various techniques of RL.    
Figure \ref{fig: RLsection} summarizes the main subsections that will be discussed in this section.

\subsection{RL and MDP Formulation: An Overview  \label{sec:RL1} }

In general, if a problem is well described as a standard MDP with stationary transition probabilities amongst the environment states, then RL will probably be effective to find an efficient solution for such a problem, not only a good solution, but also a long-term policy that maintains maximal gains over time. That does not mean one needs to fully describe the MDP (i.e. all the transition probabilities), but just that an MDP model is expected to fully characterize the environment in hand.
Conversely, if you cannot map a problem onto an MDP, then the theory behind RL makes no guarantees of any useful result. In which case, a conventional greedy technique might perform as good, or sometimes better compared to RL.

The MDP is the primary choice of modeling the problems that have a temporal structure, such as monitoring and control, which frequently arises in health monitoring applications. In these problems, the aim is to deduce a policy to be executed in real-time (at every time step). This is in contrast to optimization formulation, whose aim is to achieve an optimal design point.
Thus, MDP formulation should be considered whenever there is a decision making problem under uncertainty. Indeed, MDP and their extensions provide a general framework of acting optimally under uncertainty.  
On the contrary, if the state of the system to be optimized can be captured as a vector which is constant or varies rarely, then regular optimization formulations are preferred \cite{puterman2014markov}. 

The MDP formulation is an abstraction of the problem of data-driven control from interaction. 
It models any learning problem with three signals passing back and forth between an agent and its environment: (i) the first signal refers to the choices made by the agent (the actions), (ii) the second signal represents the basis on which the choices are made (the states), and (iii) the third one defines the agent's reward. 
Despite being this framework not sufficient to model all decision-learning problems adequately, it has proved to be widely valuable and applicable.  

An MDP is defined as the tuple $\langle\S,\A,\R,p,\gamma, \rangle$. At every time step $t$, the agent receives a representation of the environment state $S_t\in\S$. The agent executes an action $A_t\in\A$ using a policy $\pi(a|s)$ in order to receive a reward $R_t\in\R$. Then, it moves forward to the next state $S_{t+1}$. $p$ is referred to as the dynamics of the MDP. It is a probability distribution $\S \times \A \times \S \times \R \to [0,1]$ defined as $p(s',r|s,a)\doteq \Pr{S_{t}=s',R_{t}=r | S_{t=1}=s, A_{t-1}=a}$. Note that useful information can be extracted from $p$. Those are the state transitions $\Pr{S_t=s',R_t=r | S_{t-1}=s, A_{t-1}=a} = \sum_{r\in\R} p(s',r|s,a)$, as well as the reward value, i.e., 
 \begin{equation}
\E{R_t|S_{t-1}=s, A_{t-1}=a} = \sum_{r\in\R} r \sum_{s'\in\S} p(s',r|s,a).
\end{equation}  
We refer to those terminologies repeatedly throughout the paper. 

{
The total feature discounted sum of rewards until some horizon $H$ is denoted as $R_t=\sum_{t'=t}^H\gamma^{t'-t}r_t$, with the discount factor $\gamma\in[0,1]$. }
The state-action value function of a specific policy $\pi$ is defined as 
\begin{equation}
Q^\pi(s,a)=\mathbb{E}_{a\sim\pi,s'\sim\T}[R_t|s_t=s,a_t=a]. 
\end{equation} 
which summarizes the sum of the rewards resulting from taking the action $a$ in state $s$ by following the policy $\pi$. The state value function $V^{\pi}(s)=\mathbb{E}_{a\sim\pi}[Q^{\pi}(s,a)]$ assesses the quality of the state when following the policy $\pi$.



In what follows, we explore general approaches to solve a given MDP. Such solutions aim to find an optimal policy that results in the maximum reward.  

\subsection{Value-based methods \label{sec:RL2} }

The main solution of MDPs is derived from the Bellman optimality equations, which is based on dynamic programming and describes the optimal action-value of a state-action pair in terms of the next ones (recursive definition)
\begin{equation}
q_*(s,a)=\sum_{s',r}p(s',r|s,a)\big[ r+\gamma \max_{a'} q_*(s',a')\big]
\end{equation}

The equation states that the optimal action-value function for a state-action pair is expressed as the expected value of the reward from taking action $a$ in state $s$, and then taking the best possible action in the next state. This two-stages view summarizes the future expected rewards through $q_*(s',a')$, and enables the optimal policy to be easily recovered by acting greedily with respect to it:
\begin{equation}
    \pi_*=\argmax_a q_*(s,a)
\end{equation}
There are two main specific instances of value-based methods that can be utilized based on the knowledge of the environment's model. In the following subsection, we explore these two instances and discuss the model knowledge.

\subsubsection{Planning: Dynamic programming}

The main solution approach to the aforementioned MDP framework is based on exact dynamic programming, which is used when the model of the environment is known.
The model is a general concept that means anything that an agent can utilize to know how the environment will transition as a response to its actions, and what reward will it yield. Usually, this is described as a function that is referred to as the `state transition function'. If the transition model is known, the generalized policy iteration (GPI) algorithms can be employed to extract an optimal policy \cite{sutton2018reinforcement}.

GPI algorithm mainly consists of two alternating steps, a \emph{policy evaluation} step and \emph{policy improvement} step. The evaluation aims to calculate the value function (or the action-value function) of the policy being followed by the agent. In contrast, the improvement step modifies the policy in a way that is guaranteed to increase the value function.
The evaluation step might occur across multiple environment's interaction steps before an improvement step is performed. If the evaluation is done at every interaction step, GPI reduces to Value Iteration (VI) algorithm; if the evaluation is allowed to continue until convergence, GPI reduces to regular Policy Iteration (PI). In practice, VI is more widely used as it allows faster improvement of the policy.

\begin{algorithm}
\SetAlgoLined
  $V(s) \in \Re$ and $\pi(s) \in \A(s)$ arbitrarily\;
  \While{$\pi\text{-}stable$ is $false$ }{
 \While{$\Delta$ is above an accuracy threshold}{
  $v \gets V(s)$\;
  $V(s)\gets\sum_{s',r}p(s',r|s,\pi(s))[r+\gamma V(s')]$\;
  $\Delta\gets max(\Delta,|v-V(s)|)$\;
  }
  $\pi\text{-}stable \gets true$\;
   \For{each state}{
   $a_{old}\gets \pi(s)$\;
   $\pi(s)\gets \argmax_a \sum_{s',r}p(s',r|s,\pi(s))[r+\gamma V(s')]$\;
   }
  \If{$a_{old}\neq\pi(s)$}{
    $\pi\-stable \gets false$\;
}
return $\pi\approx \pi_*$
}
 \caption{Generalized Policy Iteration}
\end{algorithm}

\subsubsection{Learning: Monte-carlo and Temporal Difference}
When the environment transition model is unknown, or hard to compute or evaluate exactly, it is often useful to learn it by interaction. This is valid provided that the interaction with the environment is not expensive (samples of experience are easily and efficiently computed), which is mostly the case with simulated environments or some test-beds. In fact, leaning the state transition model is a main driver and motivation for most of the works that utilize RL. This feature has two main benefits. First, it abstracts the need to identify and design models that precisely describe the environment dynamics, and simply estimates it through samples. 
Second, in case the model is inaccurate, or in case a change occurs in the described model, the learning-based approach will adapt to this change since the approximation from samples is a continuous process. We note that the aforementioned two benefits are the main features of works that deploy RL as an optimization technique. Although there might exist other optimization approaches that might still provide the learning and adaptability features (e.g., online convex optimization \cite{hazan2019introduction}), the RL approach remain by far more popular, partly due to wide range of available resources (code-bases). An important concern is the speed of the RL adaptability, which is a topic  of recent focus in research.

Learning through samples can be done in two main ways, 1) the conventional Monte-Carlo (MC) style, or 2) the Temporal Difference (TD) methods. An MC-based method simply generates a lot of experience tuples and utilizes them to estimate the values of a state. In RL, the experience tuple refers to a state, action, next state, and reward tuple $(s,a,s',r)$. MC-based method averages the subsequent rewards from a given state across multiple episodes to get its value $v^{\pi}(s)$ (the same principle applies to state-action value function  estimation ($q^{\pi}(s)$) ).  MC  is known to be an un-biased estimator but can be slow \cite{sutton2018reinforcement}.

On the other hand, TD-based methods are direct sample-base implementation of the Bellman equations. similar to MC methods,
TD methods can learn directly from raw experience without a model of the environment's dynamics. However, TD methods update estimates based in part on other learned estimates, without waiting for the episode to terminate. This process of updating an estimate based on another one is referred to as bootstrapping. In other words, the TD methods can be realized through replacing the exact value functions in the bellman equations with samples. Bootstrapping based methods are much faster in approximating the values. However, they are biased estimators. Nonetheless, there are several bounds on the bias of that estimator, and in practice, it usually delivers very good results and is widely adopted \cite{sutton1988learning}. The best representative of TD methods, and the most adopted RL approach is referred to as Q-learning. 

Q-learning \cite{watkin} is a model free algorithm that is based on the iterative execution of Bellman equation. Specifically, the q-learning algorithm can be realized by transferring the Bellman equation into an assignment. 
\begin{multline}
    Q(S_t,A_t)\gets Q(S_t,A_t) + \alpha [R_{t+1}\\+\gamma \max_a Q(S_{t+1}, a) - Q(S_t,A_t)]
\end{multline}
Extensive theoretical analysis that guarantees the convergence can be found in  \cite{bertsekas2019reinforcement}. The iterative application of Bellman equation will eventually converge to the optimal action value function $Q^*$. Once this optimal function is generated, the optimal function can be recovered by acting greedily with respect to this function. 

Note that neural networks are often used as function approximation in evaluating the Q-function, leading to the deep Q-learning algorithm described next. While the introduction of function approximation might seem straightforward from the implementation described below, there are theoretical subtleties pertaining to the fact that convergence is no longer guaranteed. Nonetheless, various techniques have been developed to stabilize learning. Those include maintaining a replay buffer, two versions of the Q-network (Q-function), an online and a previous copy, which are used as per the algorithm.


\begin{algorithm}
    \SetKwInOut{Input}{input}
    \SetKwInOut{Output}{output}
    
    \Input{Environment simulator}
    \Output{$\theta$: The NN parameters for the approximation $Q_*$.}
\SetAlgoLined
\DontPrintSemicolon
Initialize parameters of the first network $\theta$ randomly\;
Initialize parameters of second (target) network $\bar\theta\leftarrow\theta$\;
  
  \For{episodes = 1:M}{
  Initialize state $s_0$\;
  \For{time step  t= 1:I}{
      /**Interaction with the environment**/
      Select state update action $a_t$ based on $\epsilon$-greedy policy\;
      Execute $a_t$, observe $s_{t+1}$ and $r_{t+1}$\;
      Store the tuple of experience $(s_t,a_t,s_{t+1},r_{t+1})$ in $\mathcal{D}$\;
      /**Updating the estimates**/\;
      Randomly sample a minibatch $\mathcal{F}=\big\{(s_t^{(i)},a_t^{(i)},s_{t+1}^{(i)},r_{t+1}^{(i)})\big\}_{i=1}^{|\mathcal{F}|}$ from $\mathcal{D}$\;
      Calculate Q-targets using the target network $Y^{(i)}\leftarrow r^{(i)}_{t+1}+\gamma\max\limits_{a} Q(s^{(i)}_{t+1},a;\bar\theta)$\;
      Fit $Q(s^{(i)},a^{(i)};\theta)$ to the target $Y^{(i)}$:$\theta\leftarrow\theta-\eta\nabla_{\theta} L(\theta)$\;
      Every $target$ steps, update the target network $\bar\theta\leftarrow\tau\theta+(1-\tau)\bar\theta$\;
      }
  }
 \caption{Q-learning with function approximation}
\end{algorithm}

\subsubsection{Integrated planning and value-learning}
Some RL methods aim to learn both, the model of the environment as well as the value function. This can be done through performing the update of the value function from both a real interaction tuple with the environment, and a simulated ones. The simulated ones are taken from a model that the agent learns (i.e., an approximation to the actual environment model). This type of RL might be beneficial to inject prior knowledge about the environment, or when the real interaction tuples are expensive. However, in most cases where a simulator is built, it is possible to just learn from the real samples without learning a model (model-free learning). 

\subsection{Policy-based methods \label{sec:RL3}}

As the original goal of reinforcement learning is to learn the optimal policy, a different approach is to search directly in the space of policies, rather than first to find the value function and then recover a policy out of it. This class of methods is known as the policy-gradients or policy-based methods. The policy is parametrized by a parameter vector, which is optimized through the known stochastic gradient descent (SGD) methods. In most practical instances, the direct application of the policy gradient does not yield high performance. This is due to the nature of SGD-based algorithms of being of high variance and prone to being trapped in local minima. To minimize the variance's effect, a policy gradient algorithm with a fixed baseline is proposed.  

The optimization in policy gradient methods is done on the cost function: $J(\mathbf{\theta}\doteq v_{\pi_{\mathbb{\theta}}}(s_0))$, which is the cost of starting from the initial state $s_0$, and following the parametrized policy $\pi_{\mathbf{\theta}}$ thereafter. The gradient of this function can be written as 

\begin{equation}
    (G_t - b(S_t)) \frac{\grad \pi (A_t|S_t, \mathbb{\theta_t})} {\pi (A_t|S_t, \mathbb{\theta_t})}
\end{equation}
through the policy gradient algorithm \cite{sutton2018reinforcement}. The $b$ function is any function of the state, and is called a baseline. If it is the zero function, the equation reduces to the ``reinforce'' equation. However, most of the works that utilize policy gradients combine it with a value-based method. This combination is known as Actor-critic and is explained in the next sub-section. Another popular option is the value function of the state. If this state value function is updated through bootstrapping. The resulting method is called Actor-critic. 

In general, the main benefit of policy search methods is that they offer practical means of handling very large action spaces, and even continuous spaces (infinite number of actions). This is possible because we are controlling the parameters of the policy, and not recovering it from a value function. Thus, we can choose the policy $\pi$ to be a probability distribution over the actions and learn the statistics of this distribution (e.g., the policy is a mapping from state to a normal distribution with learned mean and variance). This is different from value-based methods that learn the action-value for each of so many actions. The drawback of these policy-based methods is that they require the generation of the full episode to get a single gradient step in the policy space, which may be slow (i.e., Monte-Carlo style gradient approximation), although similar to what the TD methods introduced.

\subsubsection{Actor-critic and variants}
Actor-critic methods are policy gradients that use the state value function as a baseline ($b(s)=V(s)$) and update this function through bootstrapping \cite{lillicrap2015continuous}. In the literature, we notice that in most instances where policy search methods are used, it is indeed an actor-critic variant. Actor critic enables the online time-step wise learning as opposed to the full episode needed in conventional policy gradient algorithms. A variant of the actor-critic algorithm is shown in the following algorithm. The variant is Deep Deterministic Policy Gradient (DDPG). The main difference compared to conventional actor critic algorithms is that it maps a state directly to an action, rather that to a distribution of actions, which allows for continuous action spaces. Similar to what was presented in the previous algorithm, we adapt neural-network based function approximation to express the value function as well as the policy.

\begin{algorithm}
\caption{DDPG Template}
Randomly initialize the parameters of the Q-network $\phi$ and policy network $\theta$\;
Initialize target networks parameters with $\bar\phi \leftarrow \phi $ , $\bar\theta \leftarrow \theta$\;
\For{episodes i = 1:M}{
    Initialize state $s_0$\;
    \For{time step t= 1:I}{
        Select action $a_t=\mu_{\theta}(s)+e^{\frac{\varphi i}{M}}\varepsilon$, $\varepsilon\sim\mathcal{N}(0,0.1)$\;
        Execute action $a_t$ and observe next
        state $s_{t+1}$, reward $r_{t+1}$\;
        Store $(s,a,s_{t+1},r_{t+1})$ in replay buffer $\mathcal{D}$\;
        Sample minibatch $\mathcal{F}=\big\{(s_t^{(i)},a_t^{(i)},s_{t+1}^{(i)},r_{t+1}^{(i)})\big\}_{i=1}^{|\mathcal{F}|}$ from $\mathcal{D}$\;
        Compute targets $Y^{(i)}\leftarrow r^{(i)}_{t+1}+ \gamma Q\big(s^{(i)}_{t+1},\mu(s');\bar\theta\big)$\;
        Compute loss function $\mathcal{L}$: $\mathcal{L}(\phi, \mathcal{B})= \mathlarger{\frac{1}{|\mathcal{B}|}}\underset{(s,a,r,s',d) \in \mathcal{B}}{\mathlarger{\sum}} \Big( Q_{\phi}(s,a) - y(s',r,d)    \Big)^2 $;\
        Update Q network parameters by gradient descent:$\phi \leftarrow \phi -\eta_{\phi} \nabla_{\phi}\mathcal{L}(\phi,\mathcal{D})$\;
        Update policy network parameters by gradient ascent:
        $\theta \leftarrow \theta +\eta_{\theta}  \nabla_{\theta} \mathlarger{\frac{1}{|\mathcal{B}|}} \underset{s\in \mathcal{B}}{\mathlarger{\sum}} Q_{\phi} \big (s,\mu_{\theta}(s) \big )$\;
         Update target networks paremters $\phi_{targ} \leftarrow (1-\rho)\phi_{targ} + \rho\phi$ $\theta_{targ}\,\leftarrow (1-\rho)\theta_{targ} + \rho\theta$\;
        }
        }\end{algorithm}


\begin{table*}
\caption{Pros and cons of the main RL methods}
  \label{tab:RLcomp} 
\centering
\begin{tabular}{ m{1.5cm}|m{4cm}|m{4cm}|m{4cm}}
\hline
& Value function-based  & Policy-based methods  & Value + Policy based methods  \\\hline  
Primary algorithms & Solving BOE and recovering the policy 
& SGD on parametrized policy 
& SGD on parametrized policy + parametrized value function  
\\ \hline 
Main strength & Fast learning through TD
  & Direct optimization on the policy
  & Variations can handle continuous actions   \\\hline 
Main weakness & Biased function value
   & High variance 
  & more parameters to tune    \\ \hline 
\end{tabular}
\end{table*}

\subsection{Multi-agent Reinforcement Learning (MARL) \label{sec:RL4}} 

In this subsection, we discuss the main reasons one might consider the MARL framework to model the system at hand and point readers to the prominent algorithms in this area. In monitoring and actuating applications (vital health signs and others), a central controller usually controls and coordinates between several modalities, each of which is an element in the controller's action vector. For example, consider the controller action $\mathbf{a}={a_1,a_2,\dots,a_n}$. While algorithms discussed in the previous section can theoretically learn the best control policy, it can be seen that the size of the action space $|\mathcal{A}|$ increases exponentially with each added modality, which deems exploration among these actions infeasible. Therefore, enhancing the scalability in such scenarios is the main motive behind utilizing MARL in monitoring (health) applications. Specifically, the MARL framework does not attempt to learn one joint policy by a central controller, but rather models each of the controlled modalities as an independent agent who optimizes its policy and explores \emph{only} among its actions, while possibly coordinating with other agents in attempting to maximize the global reward.

Note that in some cases, the distribution is done by design (i.e., there exists no central controller due to security or communication concerns). Once agents are decentralized, there are several design options according to which MARL is categorized. If the reward signal is the same for all agents, the setting is known as cooperative MARL. If the rewards have to sum to a constant value, the setting is known as competitive; all other settings are referred to as mixed. We note that the study of mixed and especially competitive settings is mainly conducted in game-theoretic and economic scenarios \cite{hernandez2019survey}. In IoT applications, the cooperative scenario is the dominant one \cite{du2020survey,liu2020multi}.

In both of these settings, another design decision arises, which is whether communication between agents is possible. If so, there exist algorithms that allow agents' to utilize each others' experience through shared or communicated experience tuples \cite{hernandez2019survey}. In addition, learned communication algorithms enable agents to learn a communication protocol through exchanging messages among themselves \cite{foerster2016learning}. However, communication between agents is often not assumed in practical monitoring scenarios due to communication and privacy constraints. This leaves two main options for designing cooperative MARL without communication, namely 1) independent learning (IL), and 2) centralized training with decentralized execution (CTDE). Independent learning is the most popular option, where each agent considers all others as part of its environment. Although this approach is prevalent due to its simplicity \cite{hernandez2019survey}, it violates the stationarity requirement assumed for Markov decision processes. Specifically, the distribution of next states and rewards for each agent might change considerably, depending on what other agents are doing \cite{pmlr-v70-omidshafiei17a}. Another option is the CTDE framework, which is more common for situations where a simulator is generally available. CTDE algorithm leverages the fact that during training, agents can efficiently communicate with each other and freely exchange information (e.g., global state, action, rewards, or network parameters) since the training is mostly done in simulators. Thus, the critic can utilize the global state information (e.g., the concatenation of all agents' observations) to guide the training of the actor, which in turn takes only the local observation as an input. Then, at execution, agents only utilize policies that are based on their local observation \cite{lowe2017multi, wang2020r}. Note that during training, the centralized critic can guide the training of the policies. However, at any time, the policy can be used independently of the critic.  

There exist hybrid approaches that learn a centralized but factored Q-value function, such as QMIX \cite{rashid2018qmix}, and QTRAN \cite{son2019qtran}. These works aim to decompose a global value function into an additive, or even non linear,  decomposition of the individual value functions. However, not all coordination scenarios are easily factorizable. (i.e., we cannot always describe the global utility function by individual utility ones).


\subsection{How to select the appropriate RL model? \label{sec:RL5}}

We describe a generic way to identify the application requirements and the suitable RL model to be used. First, we summarize in Table \ref{tab:RLcomp} and Table \ref{tab:cat} the main pros and cons of diverse RL methods, as well as the categorization tool that can be used to specify the application needs, respectively. Then, according to these requirements, the appropriate solution approach with justifications are explained.  
\begin{table}
    \caption{Categorization tool}
    \label{tab:cat}
    \centering
\begin{tabular}{m{2.5cm}|m{4cm}}
\hline 
Category & Type \\\hline
Action space & \begin{itemize}
    \item Continuous
    \item Discrete
\vspace{-\baselineskip}\end{itemize}\\\hline
State space &\begin{itemize}
    \item Continuous
    \item Discrete
\vspace{-\baselineskip}\end{itemize}\\\hline
Transition model &\begin{itemize}
\item Known - Deterministic
\item Known - Stochastic
\item Unknown 
\vspace{-\baselineskip}\end{itemize}\\\hline
Sample complexity &\begin{itemize}
\item Low (simulators)
\item	Medium (simulators with complex models)
\item	High (real-world samples)
\vspace{-\baselineskip}\end{itemize} \\\hline
Objective&\begin{itemize}
 \item Discounted rewards 
 \item	Average rewards
\vspace{-\baselineskip}\end{itemize}\\\hline
\end{tabular}
\end{table}

It is assumed that the targeted application already contains a temporal structure (i.e., monitoring and/or control). Thus, we are not only interested in finding the optimal decision variables of a well-defined problem, but rather interested in finding a policy (i.e., mapping) that describes the agent behaviour in uncertain states.   
Hence, after categorizing the application's needs based on the above mentioned points, the below recommendations can be used as initial proposals for the design of an RL-powered solution:   
\begin{itemize}
    \item Discrete action/state spaces with known models (whether stochastic or deterministic ): Dynamic Programming (DP) approach (i.e., GPI or VI algorithms) are recommended since they can utilize the known dynamics to  reach speedily to the optimal policy. 
    \item Discrete, limited action space with continuous state-space: Value-based approaches (i.e., Q-learning and variants) have proved their efficiency in the literature as long as the number of actions is suitable to be cast as a Neural Network (NN) output layer.   
    \item Continuous space for both states and actions: Actor-critic approaches (i.e., DDPG and variants) are recommended since they directly optimize the continuous variables of the policy and do not need to represent each action independently, e.g., as an independent neuron in the NN output layer.   
    \item High sample complexity: Adding a  planning component is recommended, specifically, when the interaction with the environment is expensive (e.g., due to lack of a simulator). In that case, agents can build an internal model of the environment and use it to obtain approximated samples to be used along with the true ones. This approach is known as planning, and it can significantly enhance the performance with a much less number of actual interaction samples if the model is reasonably accurate.   
\end{itemize}

\subsection{Quantitative Comparison \label{sec:RL6}}

In this subsection, we empirically demonstrate the difference between the value-based methods with function approximation represented by DQN and actor-critic methods represented by the DDPG variant, which is specialized in an environment with continuous actions. The objective of these experiments is to investigate the performance of DRL in two main scenarios that can be followed in monitoring applications. In the first one, we discretize the action space according to different resolutions and then deploy DQN. In the second approach, we directly deploy the DDPG algorithm for the testing. The testing environments used are Lunar \cite{openai_gym_nodate} , a benchmark environment from open AI. Lunar simulates the control of a landing of an aerial vehicle through controlling two engines. The action vector consists of two real values vector from $-1$ to $+1$. The first controls the main engine's throttle (from off to full power), while the second value controles the orientation $-1.0$ to $-0.5$ for the left engine, $+0.5$ to $+1.0$ for the right engine, and off otherwise. We also use another specialized health monitoring environment proposed in \cite{I_SEE} whose action space also consists of vectors of two elements, each in the range $0$ to $1$, and control the transmission power and compression ratio of a secure remote health monitoring system, respectively.

Q-learning algorithms require discretization of the action space. Too dense discretization is prone to the curse of dimensionality (exponential number of actions) and leads to slower learning and convergence to policies with moderate performance. On the other hand, discretization with low resolution does converge faster, but the performance is also limited as expected since the actions are not accurate enough for the environments (i.e., engine thrust actions in the testing environments cannot be tuned freely but according to a step size).

\begin{figure}
\centering
\includegraphics[width=0.45\textwidth]{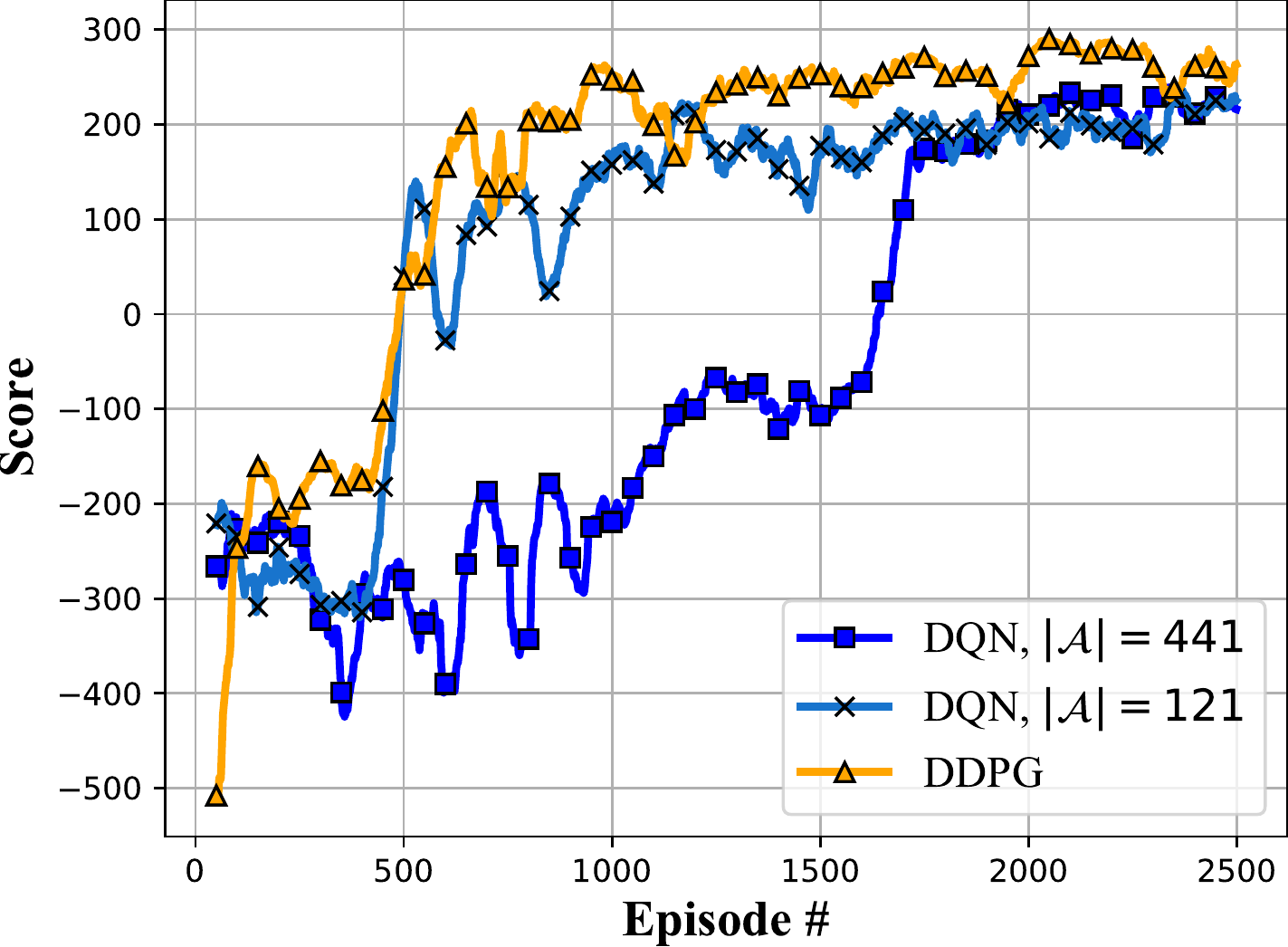}
\caption{DQN with discretization and DDPG methods performance on the Lunar benchmark environment}
\label{fig:lunar}
\end{figure}

\begin{figure}
\centering
\includegraphics[width=0.45\textwidth]{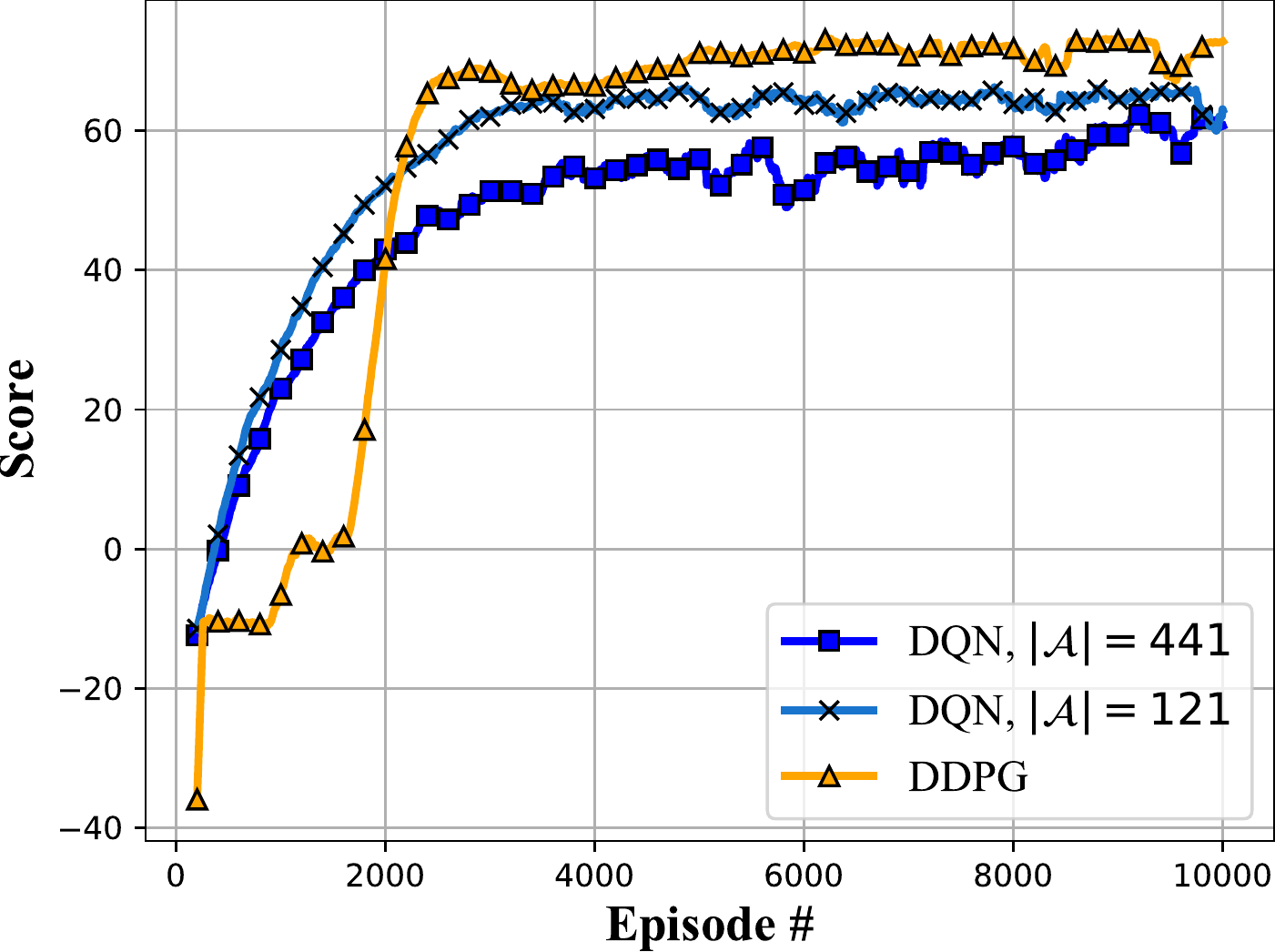}
\caption{DQN with discretization and DDPG methods performance on the custom health monitoring environment}
\label{fig:Sys2_bc}
\end{figure}

In Fig. \ref{fig:lunar}, the performance of DDPG is expectedly better as the continuous action domains are natively supported, as explained earlier. However, this comes at the cost of more parameters to tune. The used parameters are shown in table \ref{table:param_2}. The same superior performance is maintained in the health-monitoring environment in Fig. \ref{fig:Sys2_bc}. However, we notice DQN is faster to reach good action at the initial phase. This is due to a smaller range of actions in this monitoring environment (i.e., 0 to 1 as opposed to -1 to 1 in Lunar), making more actions similar to each other, and the exploration requirements less. In conclusion, these empirical experiments suggest that when the action range is limited (i.e., action precision requirements are not high), then discretization and deep Q-learning can be used to deliver satisfactory results with a relatively small number of hyperparameters to tune. On the other hand, if the precision is not known, or the action is a high dimensional vector, then DDPG can deliver higher performance but requires more parameters to tune per design.

\begin{table}[t]
\caption{DQN Parameters}
\begin{tabularx}{0.48\textwidth}{l X}
\hline
\label{table:param_2}
Parameter & Value \\~\\ \hline
Discount factor $\gamma$ & $0.9$\\
Exploration rate  $\epsilon$ & $1$, with $0.995$ Decay\\
Q-Network neurons/layer  & $24, 64, 64, |\mathcal{A}|$  \\
Q-Network learning rate  & $10^{-4}$\\
Activation function & Leaky ReLU, $0.01$ -ve slope \\
Optimizer& ADAM \cite{adam} \\
Replay buffer size $|\mathcal{D}|$  & $10^{5}$\\
Batch size $|\mathcal{F}|$ & $64$\\
Soft update factor $\tau$ & $5\times10^{-4}$\\
Soft update period  $target$ & $4$\\ \hline

\end{tabularx}
\end{table}

\begin{table}[t]
\caption{DDPG Parameters}
\begin{tabularx}{0.48\textwidth}{l X}
\hline
\label{table:param_2}
Parameter & Value \\~\\ \hline
Discount factor $\gamma$ & $0.9$\\
Exploration  & Ornstein Uhlenbeck process $\theta_{OU}=0.2$, $ \sigma=0.15$ \\
Actor-Network neurons/layer  & $24, 64, 64, 2$  \\
Critic-Network neurons/layer  & $24, 64, 64, 1$ \\
learning rates (actor, critic)  & $10^{-4}, 10^{-3}$\\
Activation function & Leaky ReLU, $0.01$ -ve slope \\
Optimizer& ADAM \cite{adam} \\
Replay buffer size $|\mathcal{D}|$  & $10^{5}$\\
Batch size $|\mathcal{F}|$ & $64$\\
Soft update factor $\tau$ & $10^{-3}$\\
Soft update period  $target$ & $1$\\ \hline  
\end{tabularx}
\end{table}

\section{ Applications and Scenarios of RL in I-Health systems \label{sec:Applications} } %

\begin{figure}[t!]
\center{\includegraphics[width=3.6in]{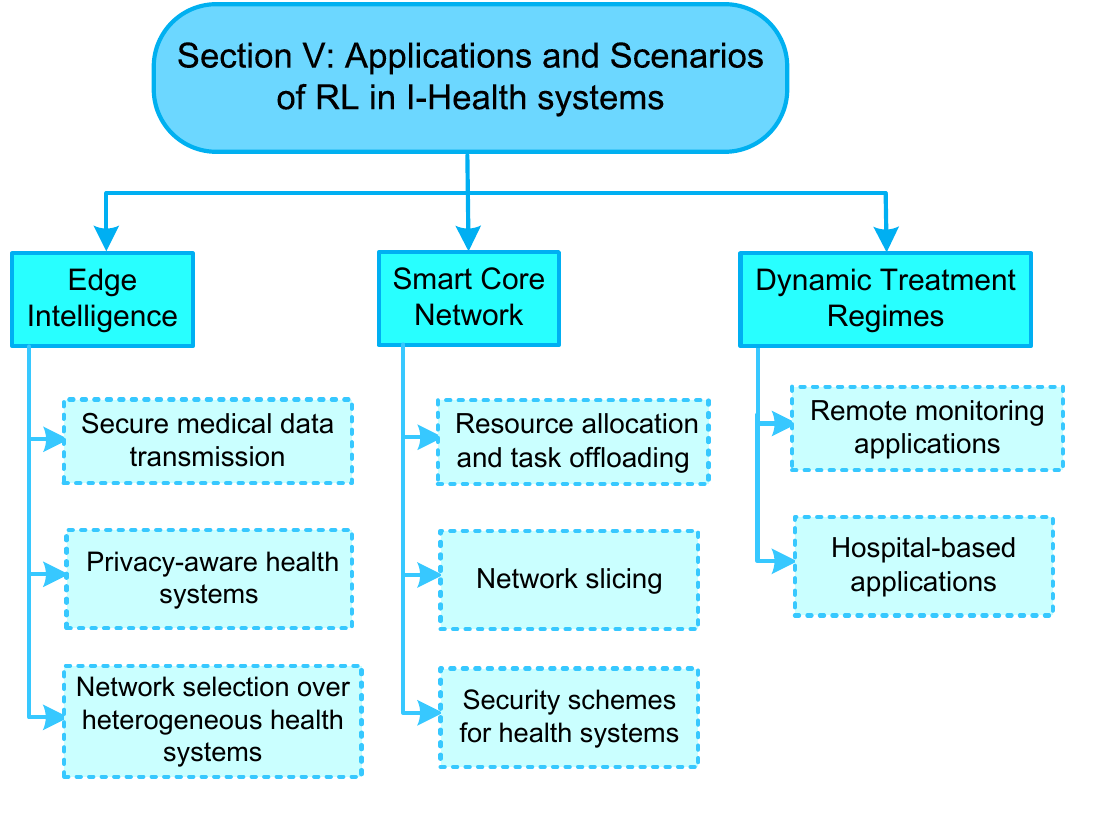}}
\caption{ Taxonomy of the applications and scenarios of RL section.  }
\label{fig:AppLsection}
\end{figure}

\begin{figure*}[t!]
\center{\includegraphics[width=6.8in]{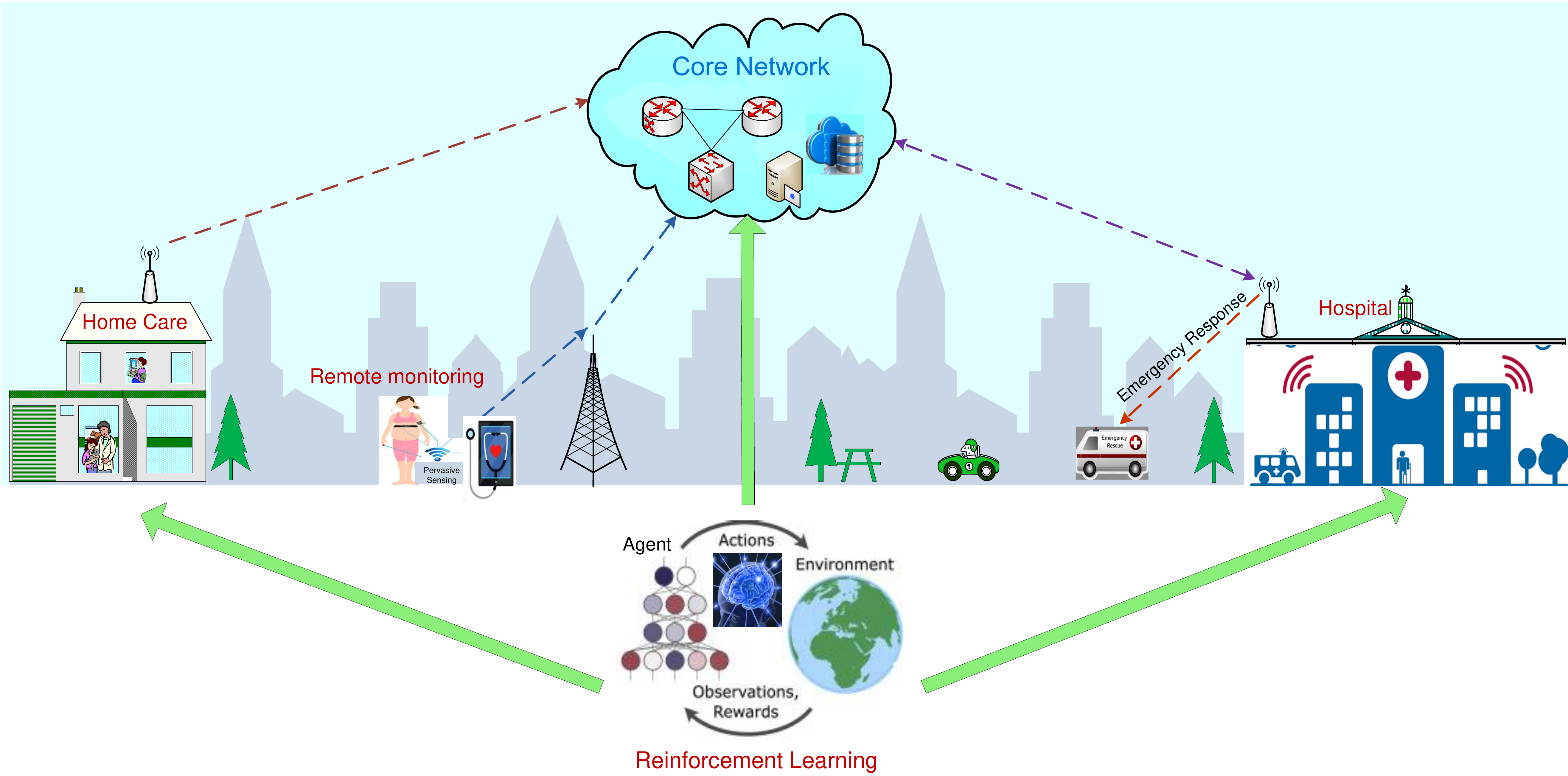}}
\caption{RL applications in I-Health system.  }
\label{fig: RL_model}
\end{figure*}

With the increasing number of data-intensive applications in healthcare domain, many challenges are emerged regarding the stringent requirements of such applications (such as ultra-low latency, high data rates, energy consumption, etc.) \cite{EdgeComputing}. Toward enabling such applications, we argue that RL can provide efficient solutions and algorithms to cope with these challenges (see Figure \ref{fig: RL_model}).  Indeed,  edge computing and smart core network architecture, as well as heterogeneous network infrastructure will play pivotal role in supporting diverse requirements of healthcare applications.  
In this section, we discuss the state-of-the-art applications of RL in three main areas related to I-health system, i.e.,  edge intelligence, smart core network, and dynamic treatment regimes.  
Figure \ref{fig:AppLsection} summarizes the main subsections that will be discussed in this section.

\subsection{Edge Intelligence  \label{sec:Edge}} 

Recently, edge intelligence (or intelligent edge) has emerged as a promising 5G technology that enables integrating edge computing capabilities with AI techniques to provide fast processing with real-time analytics. Google, IBM, and Microsoft proposed the use of intelligent edge in many applications, including agriculture precision using cognitive assistance, smart health systems \cite{MEdge-Chain}, and spanning from live video analytics \cite{kang2017neurosurgeon}. However, developing intelligent edge is still facing several challenges due to the computational and data-intensive nature of AI models. AI software, with its range of uses, demands powerful processes across computations, networking, and storage. For instance, most of the existing deep learning algorithms, e.g., convolutional neural network (CNN) and recurrent neural network (RNN), are highly resource intensive, since they were initially intended for applications, namely, natural language processing and computer vision. Indeed, the obtained high performance of deep neural networks is coming from the complex structure, the number of layers, the large amount of labeled data required for learning, and the significant resources consumed in the training phase. 

\begin{figure}[t!]
\center{\includegraphics[width=3.4in]{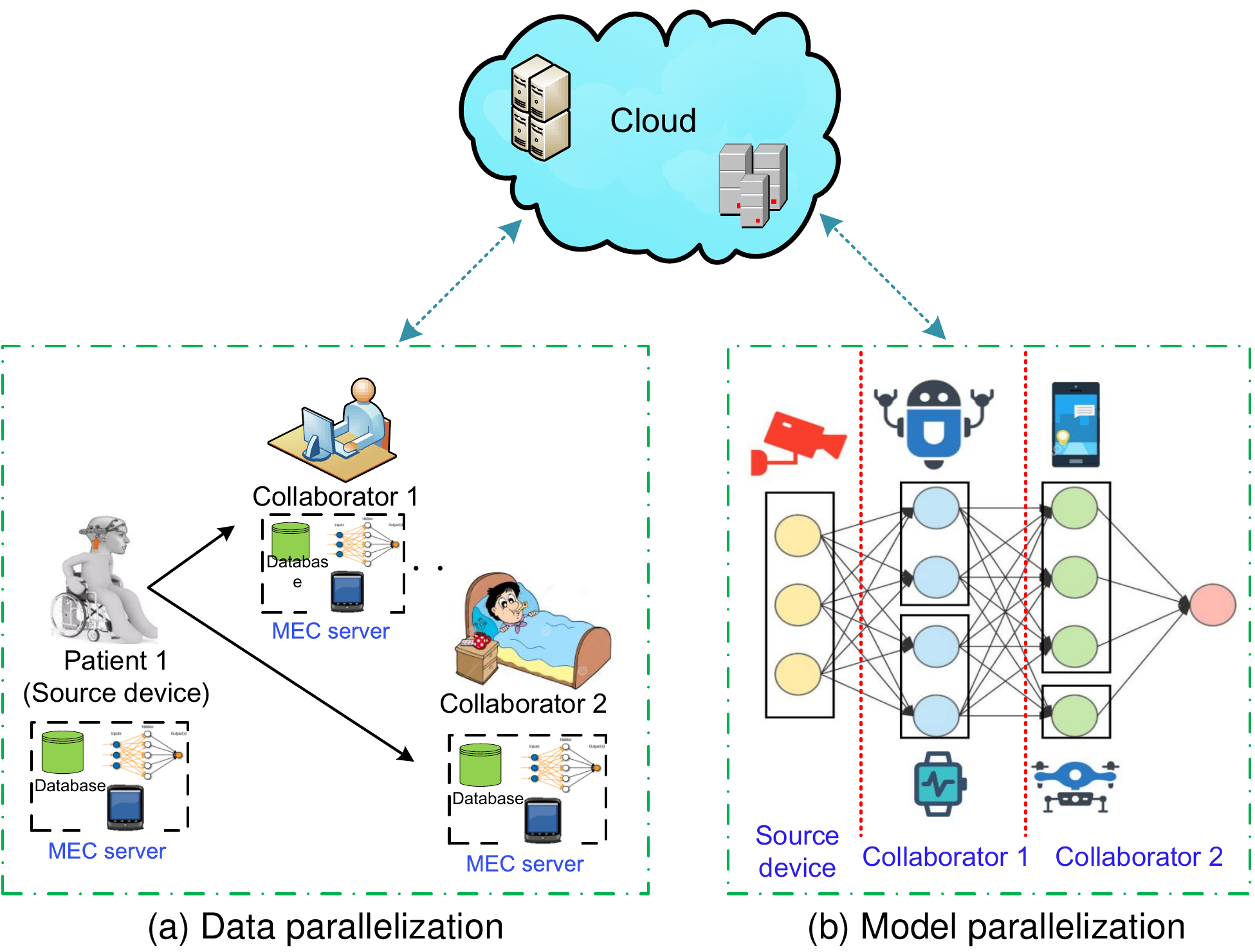}}
\caption{ Collaborative learning: data and model parallelization.  }
\label{fig:Collaborative}
\end{figure}

To decrease the overhead caused by the learning and inference processes, collaborative learning approaches have been proposed in the literature. In particular, two collaborative modes were designed (see Figure \ref{fig:Collaborative}). The first mode, namely data parallelization,  consists of distributing the training task over multiple collaborators (MEC servers). Each collaborator participates in the training phase using its private dataset and creates its individual model. Then, different collaborators exchange their parameters to design a final model. Such collaborative training, called also federated learning \cite{Clinical_Decision, naram}.  
The second mode, namely model parallelization, is related to the inference phase (i.e., execution or testing phase), where the collaborative executions are applied by dividing the complex AI models into small tasks and distribute them among the edge nodes  \cite{kang2017neurosurgeon}. Specifically, the deep neural network can be divided into segments, and each segment (i.e., one or multiple layers) is allocated to a helper. Then, each helper shares the output of its segment execution to the next participant until generating the final result/prediction.

\begin{figure*}[t!]
\center{\includegraphics[width=7.0in]{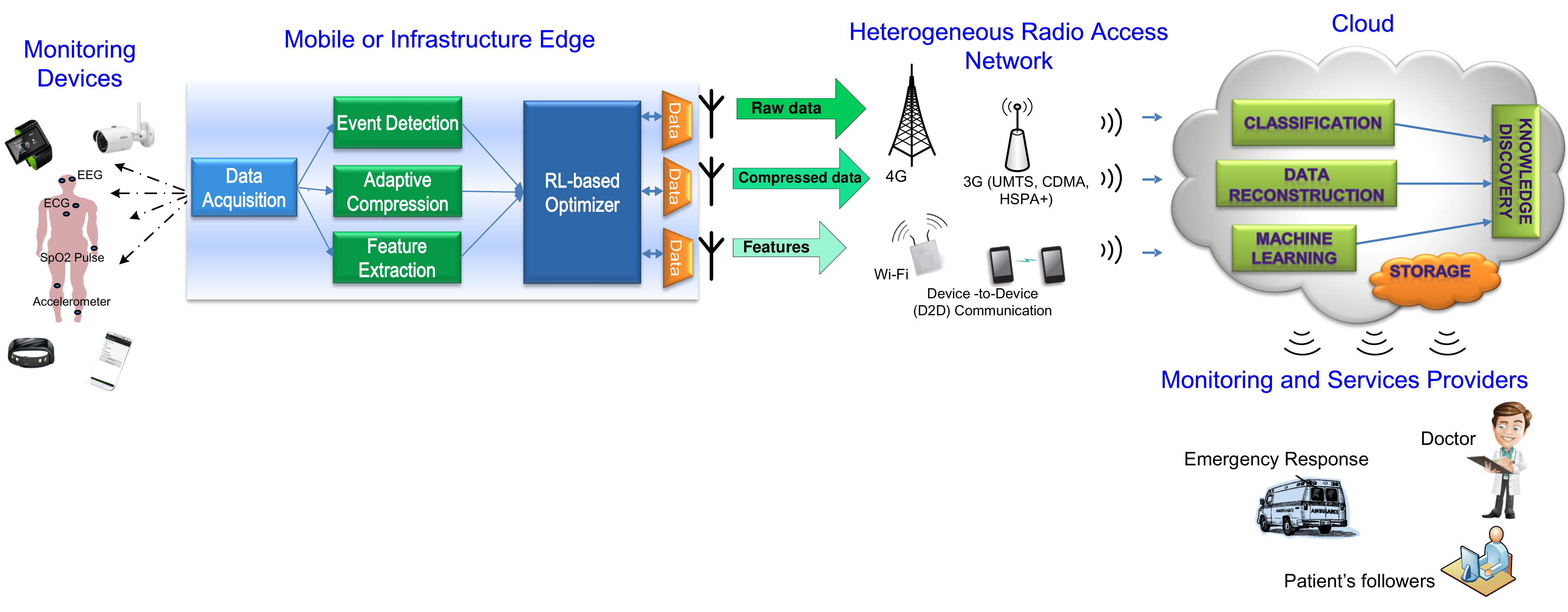}}
\caption{The implementation of edge functionality within the I-Health system.  }
\label{fig: edge}
\end{figure*}

On the other hand, bringing the edge intelligence close to the users/patients, using multi-access edge computing (MEC), along with transmitting the data over a heterogeneous 5G network is a key for supporting smart health applications \cite{EdgeComputing}. Leveraging ubiquitous sensing, heterogeneous network, and intelligent edges, the proposed MEC platform in Figure \ref{fig: edge} can real-timely monitor people's daily life and provide intelligent healthcare services, such as emergency detection and notification, epidemic prediction, and occupational safety.  
In Figure \ref{fig: edge}, the edge node can: 
\begin{enumerate}
	\item acquire the medical and non-medical data from various monitoring devices;
	\item implement advanced AI techniques using the acquired data for data compression, event-detection, and emergency notification; 
	\item run RL-based solution for optimizing medical data delivery; and 
	\item forward the paramount data or extracted features of interest to the cloud/healthcare service providers.  
\end{enumerate}
Interestingly, various health-related applications (apps) can be implemented at the edge level for supporting real-time patient-doctors' interactions. Hence, the patients can participate in their treatment regimes while interacting with their doctors anywhere and anytime. Also, specialized context-aware processing schemes can be implemented near the patients to allow for optimizing medical data processing and transmission based on the context (i.e., data type, supported application, and patient's state) and wireless networks' dynamics \cite{EdgeHealth, EEG_Transceiver}. 

In what follows, we will focus on some use cases that implement RL-based solutions at the network edge for enabling efficient healthcare services.   

\subsubsection{ Secure medical data transmission } 
\begin{figure}[t!]
\center{\includegraphics[width=3.25in]{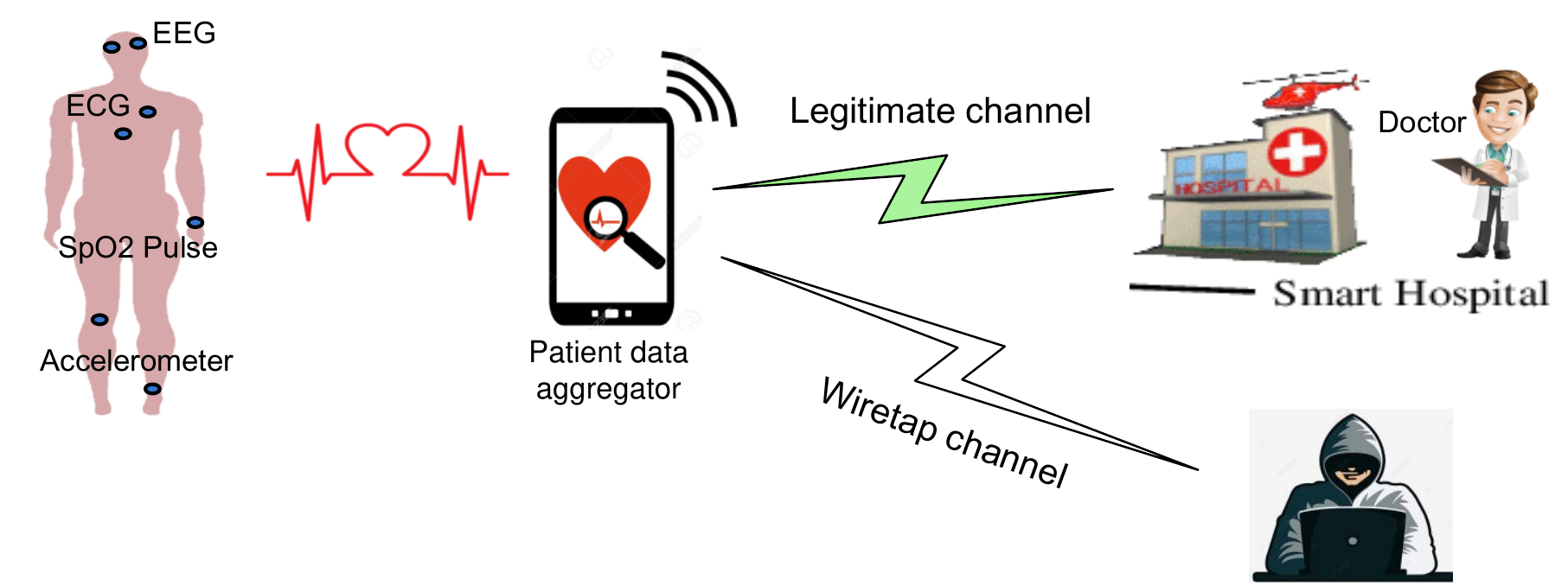}}
\caption{PLS system model for securing medical data transmission.  }
\label{fig:PLS}
\end{figure}

Secure medical data transmission from a patient to healthcare service providers is mandatory for smart healthcare systems. Traditionally, for supporting secure wireless communication, different security schemes have been implemented either at the upper layers of the Open System Interconnection (OSI) model using cryptographic techniques, or leveraging Physical-Layer Security (PLS). The latter refers to the fact that data transmission in a wireless channel typically is subjected to random channels' effects, e.g., noise, attenuation and multi-path fading. PLS leverages such channels' effects to
secure the wireless transmission by utilizing the secrecy capacity notion \cite{I_SEE}.  Indeed, secrecy capacity refers to the secure wireless communication rate that can be transmitted through the legitimate channel without being wiretapped by the eavesdropper (see Figure \ref{fig:PLS}). 

Q-learning and DRL schemes have been recently utilized for PLS to develop a learning-based model for secure medical data transmission in dynamic wireless environment  \cite{7398138}. For instance, the authors in \cite{8686092} exploit stochastic game, i.e., considers transmitter, receiver and multiple attackers as players, to tackle the PLS problem, while leveraging Q-Learning algorithm to control the power allocation and enhance secrecy capacity. 
In \cite{RL_PLS2021}, the authors leverage RL to optimize the authentication
policy for controller area networks' bus authentication. In particular, the proposed solution uses RL to select the authentication mode that captures the physical-layer features (i.e., the arrival intervals and signal voltages of the received messages), while detecting spoofing attacks and improving authentication accuracy.  
In \cite{DRL_PLS2019}, a physical-layer anti-eavesdropping scheme is proposed for multiple-input-single-output visible light communication  wiretap channel. Indeed, a DRL-based smart beamforming solution is proposed to reduce the eavesdropped signal level while enhancing the received signal level at the legitimate receiver. Also, an actor-critic algorithm is used to improve the learning rate, while utilizing the information of the high-dimensional structure of the beamforming policy domain. 
In \cite{I_SEE}, the authors exploit RL to provide secure and optimized medical data delivery over highly dynamic healthcare environment. Indeed, they show that RL-based solution could significantly reduce the computational overhead resulting from re-optimizing the solution when changes happen, compared to the optimization-based solutions in  \cite{bilal2019, Tacticallutif}.  

\subsubsection{Privacy-aware health systems}
MEC architecture can reduce the energy consumption at the IoMT devices by processing the acquired information at the edge nodes (e.g., base stations, access point, and laptops) that have sufficient computational and energy resources. However, offloading such private data from the IoMT devices to the edge nodes comes at the risk of violating privacy conditions. An eavesdropper, near to the IoT devices, can detect the location and data pattern of the patients during data offloading. Thus, efficient privacy-preserving schemes should be implemented near to the patients.  

In this context, the authors in \cite{LearningBasedPrivacyAware}  present a privacy-aware offloading scheme that aims at obtaining the offloading rate of the acquired medical data in energy-harvesting powered healthcare systems. Specifically, RL-based algorithm is utilized to define the optimal offloading policy while accounting for the energy harvesting, radio channel states, size and priority of the acquired data, as well as the battery level. Moreover, a transfer learning mechanism,  post-decision state (PDS) method, and Dyna architecture are used to accelerate the learning process.  
The problem of decentralized real-time sequential clinical decisions problem has been tackled in \cite{Clinical_Decision}, considering MEC architecture. In particular, a Double Deep Q-Network (DDQN) algorithm is implemented, at each edge node, to define the optimal clinical treatment policy, while using decentralized federated framework for DDQN models aggregation from all edge nodes. Moreover, two additively homomorphic encryption schemes have been developed to guarantee the privacy of the acquired medical data during the training process.  
 A lightweight privacy-preserving scheme is presented in \cite{OIDPR},  based on  Q-learning. In particular, the additive secret sharing and edge computing 
are used to enhance the efficiency of data encryption for diabetic patients. Indeed, the edge servers are used to decrease the needed model's updates and drug dose calculation times. Hence, the proposed scheme allows for reducing the demand for computing power, while maintaining the required efficiency and privacy protection.  

\subsubsection{Network selection over heterogeneous health systems} 
\begin{figure}[t!]
\center{\includegraphics[width=3.2in]{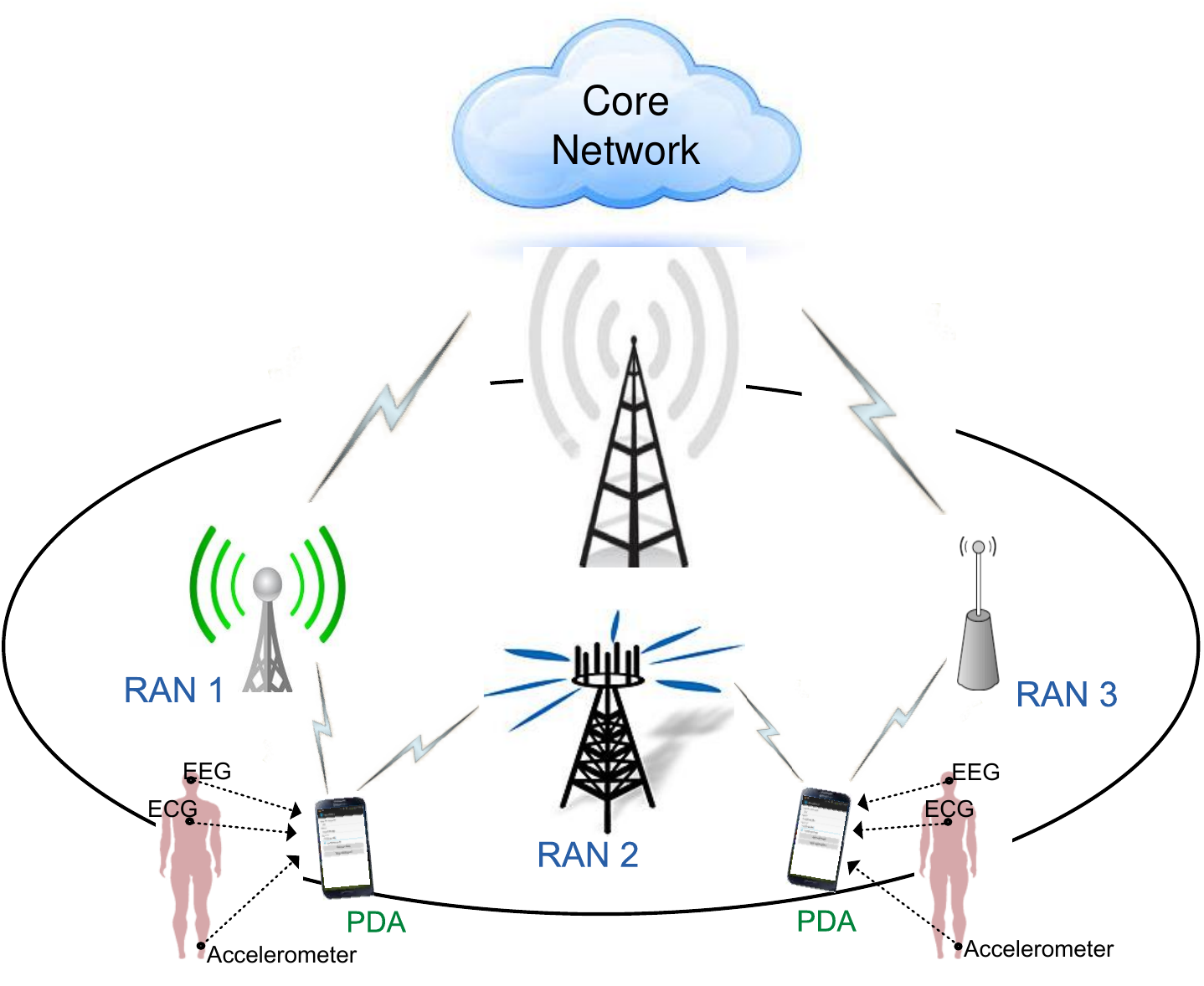}}
\caption{Heterogeneous I-Health system with multi-RAT.  }
\label{fig: RAN_selec}
\end{figure}

5G HetNet is considered as an important feature that can provide high spectral efficiency, low latency, and high throughput for diverse types of applications. Thanks to the availability of several cellular, D2D communication \cite{Network_D2D}, WiFi, and fixed access technologies, the performance of I-Health system can be significantly enhanced by enabling data transfer from edge nodes to the core network in an energy-efficient manner, while maintaining strict QoS requirements (see Figure \ref{fig: RAN_selec}).  
However, this calls for designing efficient network selection/association algorithms to allow for connecting different users to the optimal network(s), while satisfying user preferences and reducing the access failure rate. Indeed, developing low-latency, highly reliable, and cost-effective network association schemes can significantly help in fulfilling diverse requirements of the I-Health systems, which include: 
\begin{enumerate}
    \item fulfilling the explosive traffic demand in remote monitoring applications,
    \item providing ubiquitous connectivity and smooth experience for different patients, and
    \item supporting ultra-low latency applications, particularly in case of adverse patient's events. 
\end{enumerate}

By using a Patient Edge Node (PEN), e.g., smartphone, equipped with different Radio Access Technologies (RATs), a patient (or user) would be able to forward its data to the healthcare service providers through multi-RAT \cite{network_system}. However, the real-time process of obtaining which RAT (or RAN) to use is not an easy task, given the RANs' dynamics and traffic variations.   
In this context, RL has been used to allow each user to learn from his/her previous experience and define the optimal RAN-selection policy \cite{RL_selec2019}. 
RL-based network selection solutions in \cite{reinforcement_NS2_dynamics,reinforcement_NS, DRL_network} have depicted the efficiency of RL in achieving swift convergence and near-optimal performance. 
Specifically, the authors in \cite{DRL_network} present a DRL-based solution that leverages dense HetNet within the smart health system to enhance network capacity, provide seamless connectivity for healthcare applications, and fulfill diverse remote monitoring applications' demand.   
First, a multi-objective optimization problem has been formulated to minimize, for each PEN $i$ over RAN $j$, the transmission energy consumption $E_{ij}$ \cite{5982690},  monetary cost $C_{ij}$, latency $L_ij$, and signal distortion $D_i$, while maintaining the application's QoS requirements. 
Given that there are $N$ PENs would transfer their medical data to the health cloud through the available $R$ RANs, the objective of the formulated optimization problem is to define the optimal compression ratio and selected RAN(s) for each PEN $i$. Hence, the RAN-selection problem in \cite{DRL_network} is formulated as: 
\begin{eqnarray}
    & \min_{  P_{ij}, \kappa_i} \quad
       \sum_{i=1}^{N}{(\sum_{j=1}^{R}{P_{ij}U_{ij}+\delta_i D_i})} \label{optimize_prob}
     \\
    & \text{such that } \qquad \qquad \qquad \qquad \notag \\
    &\sum_{i=1}^{N}{\frac{P_{ij}b_{i}}{r_{ij}}}\leq T_j , \ \forall j=1..R\\
    & \sum_{j=1}^{M}P_{ij} =  1, \ \forall i=1..N  \label{C2}  \\
    &  0\leq P_{ij} \leq 1, \\
    & 0< \kappa_i < 1, 
\end{eqnarray}
where
\begin{align}
    U_{ij} = \alpha_i  E_{ij} + \beta_i C_{ij} + \gamma_i L_{ij}. 
\label{eaut}
\end{align}
Second, the RAN-selection problem is solved using DRL. In particular, the set of states is formulated as $S=\{T_1^r,T_2^r,..T_R^r, PEN_i\}$,  where  $T_j^r$ refers to the available time slots at RAN $j$. The set of actions $A$ is defined as $A=\{P_{i,1},P_{i,2},..P_{i,R}, k_{i}\}$, where  $P_{ij}$ is the RAN indicator that refers to the fraction of data of PEN $i$ that will be sent through RAN $j$, and $T_{j}$ is the available fraction of time over RAN $j$ that can be used by different PENs (i.e., resource share).  Then, due to the continuous action space of the formulated problem, a DDPG algorithm is proposed to solve the formulated probelm.  

The existence work for network selection has been also extended for large scale systems by leveraging MARL techniques. Given that multiple users are competing on the available resources over different RANs,  MARL can provide a decentralized solution that optimizes the individual users' policies.  
For instance, the authors in \cite{MARL_NS2_smart} present an optimized solution, leveraging MARL techniques, for maximizing the average system throughput over multi-RAT network, while maintaining the users' QoS constraints. However, this work ignores the energy consumption constraint and the high-level applications' requirements, such as the reliability, which opens the door for future research directions.  

\subsection{Smart Core Network  \label{sec:Core}}  

RL has been extensively used  as an efficient tool for optimizing core network functions and dynamic treatment regimes at the cloud. In particular, with the rapid evolution of I-heath systems along with cloud computing, RL techniques gained much interest for optimizing policy-making and treatment strategies.  
In what follows, we present some use cases that consider RL techniques at the core network for optimizing the performance of I-health system.   

\subsubsection{Resource allocation and task offloading}
\begin{figure}[t!]
\center{\includegraphics[width=3.1in]{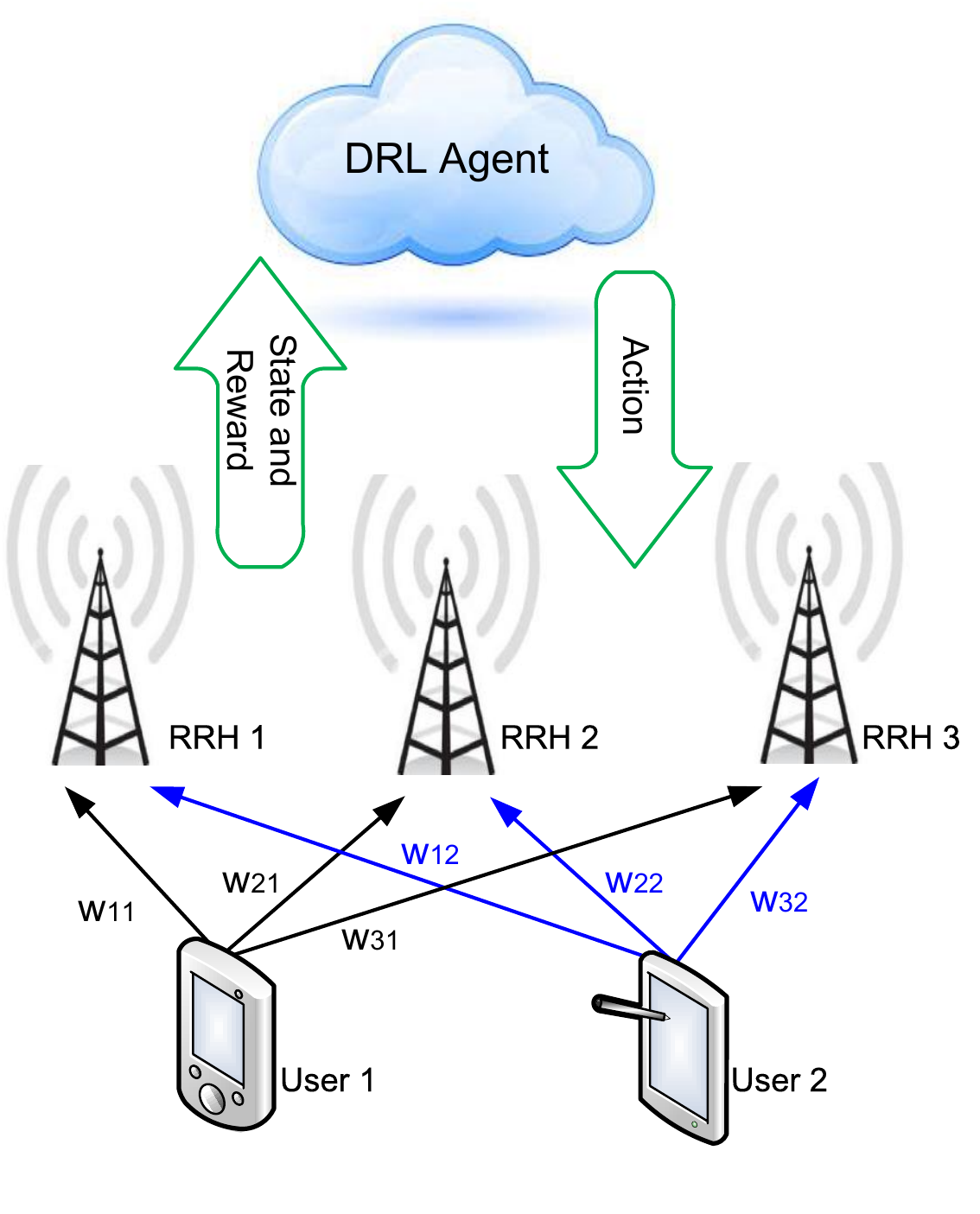}}
\caption{RL-based dynamic resource allocation solution for cloud RAN architecture.  }
\label{fig:resourceAllocation}
\end{figure}

RL is considered as one of the promising solutions for dynamic resource allocation in wireless communication systems \cite{r1, rarab, a13}. The effectiveness of DRL-based solutions for dynamic resource allocation are twofold. First, the obtained actions (or decisions) are taken with the help of DNN, which provides accurate solutions for the whole operational period \cite{silver2016mastering}. Second, leveraging DRL allows for considering the state transition overheads, e.g., power consumption related to the transition from one transmission mode to another. Such transition states have been shown to be significant \cite {xu2015automated}, however they are neglected in most of the resource allocation schemes that provide optimal/suboptimal solutions for the current time slot only.   
In this context, the authors in \cite {r1} have leveraged the power of DRL 
in solving complex control problems to develop a DRL-based solution for dynamic resource allocation in a cloud-RAN architecture (see Figure \ref{fig:resourceAllocation}). The proposed framework aims at minimizing the total power consumption while fulfilling the requested demand for different users. In particular, the state space is defined as a function of the binary states (active or sleep) of each Remote Radio Heads (RRHs) and the required demand of each user. The action space refers to which RRH will be turned on or off, while the reward function is defined as a function of  the maximum possible value of the total power consumption. Then, the DNN is used to approximate the action-value function, while formulating the resource allocation problem (in each decision epoch) as a convex optimization problem, as follows:
\begin{eqnarray}
    & \min_{  w_r, u} \quad
       \sum_{r\in A}\sum_{ u \in U}{|w_{r,u}|^2} \label{optimize_prob1}
     \\ 
    & \text{subject to } \qquad \qquad \qquad \qquad \notag \\
    & SINR_u\leq \gamma_u, u \in U \\
    &\sum_{u \in U}{|w_{r,u}|^2}\leq P_r , \ \forall r\in A\ \\
    & \gamma_u= \alpha_m(2^{R_u/B}-1), 
\end{eqnarray}
where $P_r$ is the maximum transmit power of RRH $r$ and $R_u$ is the requested demand of user $u$. The optimization variables in (\ref{optimize_prob1}) are $w_{r,u}$, which refer to the beamforming weights. 

In \cite{rarab}, a  centralized DRL-based downlink power allocation scheme is proposed for a multi-cell system with the aim of maximizing the total network throughput. 
In particular, a DQN approach is applied to define a near-optimal power allocation policy. The obtained simulation results depict that the proposed scheme outperforms the conventional power allocation schemes in a multi-cell scenario.  
In \cite{a14} the authors develop a distributed DQN-based spectrum sharing approach for primary and secondary users in a non-cooperative fashion. The primary users use a fixed power control strategy while the secondary users learning autonomously to adjust the transmission power following the obtained policy. In \cite{a015}, the authors use the DRL approach to perform joint user association and resource allocation (UARA) over  heterogeneous network. The ultimate goal is to maximize the long-term utility of the network while ensuring QoS requirements. In \cite{a16}, the authors use a multi-agent DQN approach to allocate power in wireless networks. The principal objective was to maximize the weighted sum-rate of the system. In \cite{a17}, the authors develop several DRL architectures, such as REINFORCE, DQN, and DDPG, for  optimizing the power allocation in multi-user cellular networks while  maximizing the overall sum-rate of the network. In \cite{a18}, the authors leverage  DRL for solving the problem of dynamic spectrum access in wireless networks. The main goal of this work was to maximize the  utility of each user in a distributed manner, i.e.,  without exchanging information.  

In \cite{8036259}, the authors present a DRL-based link scheduling algorithm to mitigate the interference impact in wireless networks.  The proposed algorithm applies MDP  to model the network's dynamics, i.e., channel state information (CSI) and cache states, at the transmitter side. In particular, a DNN model is utilized to learn  the states' variations.   
In \cite{8119495}, DRL is exploited to address the mobility impact in wireless networks, where a  recurrent NN and a convolutional NN  are used to extract the features from the received signal strength indicators. 
In \cite{7997286}, a  power consumption minimization model is proposed, where a DRL algorithm is used to control the activation of remote radio heads (RRH), while considering the power consumption caused by the RRH on-off state transition.  In \cite{8362276},  a DRL-based model is presented for cache management, where the requests' frequencies from diverse users are considered as the input state, while defining the actions based on whether to cache the requested content or not. 

Mobile Cloud Computing (MCC) has also gained much interest recently to enable pervasive healthcare services in an energy-efficient and cost-effective way \cite{Softwaredefinhealth, MCCoffloading, UbeHealth, FedHome}. The main objective of MCC is to efficiently offload computation-intensive tasks from mobile devices (i.e., battery-operated devices) to the cloud, while guaranteeing continuous and uninterrupted healthcare services.
For instance,  a model-free RL-based scheduling approach, leveraging Q-learning, has been proposed in \cite{MCCoffloading} to enhance the task scheduling and dynamic computation offloading from mobile devices to the cloud. Indeed, this work focuses on three main objectives for optimizing computation-intensive tasks offloading, i.e., battery lifetime, diagnostic accuracy, and processing latency. 
The work in \cite{UbeHealth} has considered the four layers, i.e., Mobile, Cloudlet, Network, and Cloud layers, to improve the network quality for healthcare services.   
Specifically, it proposes a network traffic analysis and prediction scheme that exploits deep learning, big data, and high-performance computing resources at the cloud for predicting the network traffic. Then, the Cloudlets and network layer leverage the predicted traffic to optimize data rates, data caching, and routing decisions, which enables the network layer to satisfy the communication requirements of the healthcare applications.

Several research studies have also tackled the human activity recognition in the field of smart health. Indeed, different practical applications, such as gym physical activity recognition and fall detection have been investigated. AI schemes play a major role in such studies through automatically detecting and identifying human activities and behavior patterns by analyzing the acquired data from various IoMT devices. The recent studies that tackled the human activity recognition problems can be categorized into two groups: ambient sensor-based and wearable sensor-based schemes. The former refers to the techniques that utilize surveillance camera, temperature, and other indoor sensors to acquire the environment-related data to detect people's daily activities within smart-assisted environment, such as smart homes and recovery centers \cite{6780615, 7589110}. The latter refers to the techniques that leverage wearable devices or smartphones to acquire and monitor vital signs or on-body physiological signals using accelerometer, magnetometer, and gyroscope sensors \cite{9409613}.  
For instance, the authors in \cite{HumActivRecog} present a semi-supervised deep learning framework that incorporates an auto-labeling technique with a long short-term memory (LSTM)-based classification scheme. The main goal of this framework is to efficiently leverage the large amount of weakly labeled data to train a classifier in order to enhance the classification accuracy of human activity recognition applications. Indeed, the proposed auto-labeling technique is developed based on DQN to address the problem of inadequately labeled data and enhance the learning accuracy. 

\subsubsection{Network slicing}

\begin{figure}[t!]
\center{\includegraphics[width=3.4in]{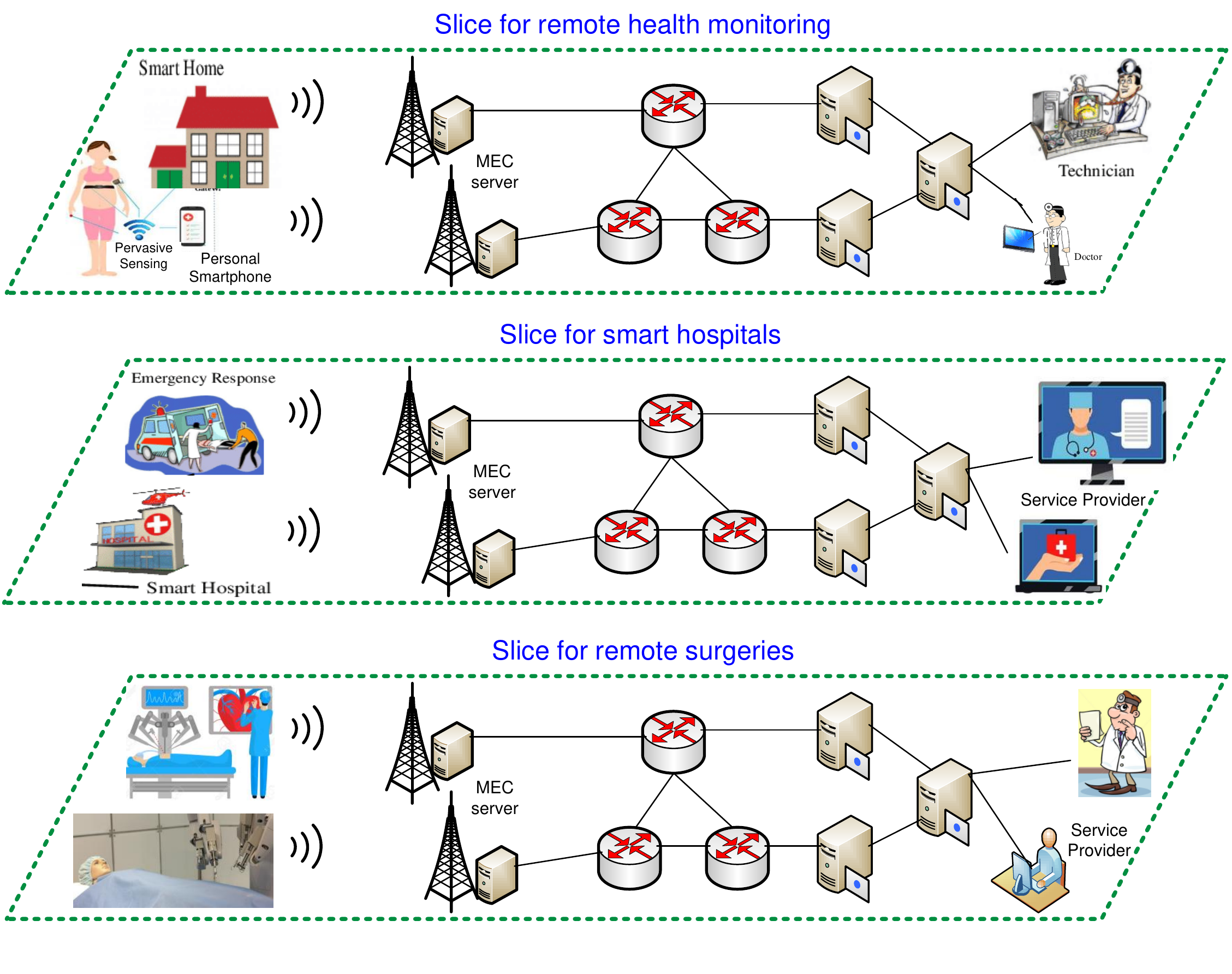}}
\caption{Network slicing for I-Health systems.  }
\label{fig: slicing}
\end{figure}

Network slicing is one of the major technologies in 5G networks that aims at enabling the realization of the growth in the Industrial IoT market. It refers to the ability of the network operators to split the physical network components into multiple logical slices (i.e. sub-networks) with different characteristics. Hence, diverse  IoT applications with different demands can be supported using  various dedicated network slices \cite{9382385}.  
Recently, network slice as a service has been emerged as a new concept that allows network operators to create a customised network slice for each user/application as a service. In the context of I-health systems, network slicing can help in: 
\begin{itemize}
    \item maximizing resource utilization by dynamically adapting available resources for each slice \cite{3434},   
    \item enhancing the system scalability while fulfilling diverse QoS requirements, and
    \item  providing high security and high privacy for sensitive healthcare applications \cite{8787, 8888, 8989}.   
\end{itemize}
Specifically, remote monitoring applications, smart hospitals, and remote surgeries demand for high data rates, supreme reliability, and ultra-low latency communication, which can be supported by novel 5G technologies such as network slicing \cite{8787}. Indeed, each health-related application can be allocated a customised network slice to support its distinctive requirements (see Figure \ref{fig: slicing}).   
Thus, the authors in \cite{124124} have investigated the potential of network slicing for supporting efficient connection between smart wearable devices  and the cloud. In  \cite{209209}, network slicing and 5G network have been leveraged for enabling remote surgeries applications with ultra-reliable requirements, ultra-low latency, and guaranteed bandwidth allocation.  

Dynamic slice allocation and dynamic resource allocation between slices are two main approaches for enhancing dynamicity of heterogeneous IoT systems, such as I-health systems. In this context, learning techniques, such as deep learning and RL, can significantly help in predicting users' demands, hence adjusting the resource allocation dynamically between different slices based on  the predicted demands \cite{2929}. 
Conventional optimization models for network slicing and queueing theoretic modeling turn to be intractable, especially for highly dynamic environments with strict reliability, bandwidth, and latency requirements.  
Thus, leveraging deep learning and DRL for dynamic network slicing and resource management in 5G network has gained much research interest recently \cite{liu2019network, DeepSlicing, 2929}.  DRL has proved its efficiency in enhancing the overall system performance in terms of throughput, latency, and reliability. For instance, the formulated network slicing problem in \cite{DeepSlicing} has been decomposed into a master problem and several slave problems. The master problem is solved using convex optimization, while DDPG algorithm is used to solve the slave problems. The main motivation of using DDPG algorithm is to learn the optimal policy for allocating the users' resources, even if there is no closed-form expression for the users' utility functions in the slave problems. 
In \cite{koo2019deep}, the authors leverage DRL to simultaneously consider varying traffic
arrival rates, dynamic applications' requirements, and limited resources availability. In particular, the network slicing resource allocation problem is mathematically formulated as a MDP, then a policy-gradient method, using REINFORCE algorithm \cite{silver2017mastering},  is utilized to solve this problem and learn the optimal policy for network slicing resource allocation.   
The network slicing resource trading process between a slice provider and various
tenants is also formulated as an MDP in \cite{kim2019reinforcement}. In particular, the MDP state space is defined as a function of the QoS satisfaction parameters. Then,  a Q-learning-based dynamic resource allocation strategy is used to maximize tenants' profit, while fulfilling diverse QoS requirements of end users in each slice. 

Interestingly, strict security and privacy requirements for sensitive IoT applications, such as I-health, can be maintained by exploiting the concept of slice isolation \cite{3535, 9369546, 9076479}.  For instance, fog computing and network slicing have been used in \cite{3535} to develop a secure service-oriented authentication scheme for IoT systems over 5G network. The main idea of this work is to isolate users/devices that have suspicious behaviour in quarantine slices until executing the necessary actions.  In \cite{9369546}, the authors integrate the network slicing concept with blockchain technology in order to support privacy isolation for a hospital network.  
Thus, we envision that implementing slice isolation concept within the I-health system can be beneficial in addressing several security issues.   

\subsubsection{ Security schemes for health systems}

Security is one of the fundamental concerns in any healthcare system. 
Typically, in I-health systems, the acquired medical data from the patients (e.g., blood pressure, blood sugar, etc.) are forwarded to the cloud servers for processing and storage, where sophisticated data mining and AI models are implemented to provide pre-diagnosis decisions. The healthcare service providers are then notified with these decisions in order to take actions and provide the appropriate responses. 
However, enabling secure data exchange, remote monitoring, data processing, and real-time diagnosis services without revealing patients' information and privacy is still challenging \cite{PP_CLOUD2019, PP_MultiSourceCloud}.   
Indeed, classical healthcare systems usually rely on  weak security schemes to protect their data processing and management, which results in serious security problems in these systems. For example, number of recorded security attacks in healthcare systems increased between 2016 and 2017 by $89\%$ as reported in \cite{ransomware}. Moreover, many reports depicted that a wave of cyberattacks have threaten several hospitals during the recent COVID-19 pandemic \cite{Cybersattack2020}. These attacks can shut down hospitals' services and hinder healthcare facilities \cite{Cybersecurity2020}. 
For example, an attacker can steel user identification by obtaining login credentials via the use of phishing emails. This attacker who has access to the medical system can falsify patients' reports as well as attacked the control system. 

Different studies have tackled such problems using RL schemes in order to protect patients' personal information and healthcare systems.   
For instance, the authors in \cite{nguyen2019deep} present a mathematical model  for  cyber state dynamics as:
\begin{align}
    & x(t)=f(t,x,u,w;\theta(t,a,d)) \\
    & x(t_0)=x_0, 
\end{align}
where the physical layer disturbances are denoted by $w$, cyber attack is denoted by $a$, and the defence strategy is denoted by $d$. The  attack-defense dynamics are presented by $\theta(t, a, d)$ at time $t$, and $x$ defines the physical state. Then, an actor-critic DRL algorithm is used to present the defence strategy against the hackers. The obtained results depict the efficiency of the proposed DRL model with its ability to obtain the optimal strategy that could enhance the system performance in terms of attacks detection  \cite{nguyen2019deep}.    
The authors in  \cite{PrivPresRL2021} develop a privacy-preserving RL-based framework for patient-centric dynamic treatment regime. This framework has been deployed at the cloud computing environment for providing: 
(i) secure multi-source data storage, 
(ii) secure non-integer processing for multiple encrypted domains, and (iii) secure RL training without leaking the data of the patients. 
In \cite{s81}, DDQN and Asynchronous Advantage Actor-Critic (A3C) algorithms have been used to solve the robustness guided falsification problem of  Cyber-Physical System (CPS). This work has shown that using traditional methods such as simulated annealing \cite{s82} and cross entropy \cite{s83} are inefficient due to the infinite state space of CPS models. On the contrary,  formulating the problem as an RL problem could obtain better results compared with the existing state-of-the-art techniques for detecting any  falsified inputs of CPS.


Table \ref{tab:security} summarizes some of the existing DRL models for network security,  which can be easily extended to protect healthcare systems.  
\begin{table*}[!h] 
	\centering
		\caption{Summary of the important studies that considered RL-based security schemes.}
	  \label{tab:security}
	  \begin{tabular}{|l|l|l|l|l|l|l|}
\hline
Ref                & Technique                                                                              & Objectives                                                                                                                          & \begin{tabular}[c]{@{}l@{}}DRL\\  algorithm\end{tabular}         & States                                                                                                                      & Actions                                                                                                                                          & Rewards                                                                                                                                            \\ \hline
\cite{s133}     & \begin{tabular}[c]{@{}l@{}} Secure data  \\ offloading \end{tabular} & \begin{tabular}[c]{@{}l@{}} Create a policy  for \\  a secure data \\  offloading\\  to avoid jamming \\ attacks \end{tabular} & DQN             & \begin{tabular}[c]{@{}l@{}} User density, \\ battery levels\\ channel bandwidth\\  and jamming \\ strength\end{tabular}  & \begin{tabular}[c]{@{}l@{}} selected edge \\ node, transmission \\  power, \\  offloading rate \\ and time \end{tabular} & \begin{tabular}[c]{@{}l@{}} Defined based on  \\ secrecy rate,  \\ communication  \\ efficiency, and  \\  power consumption \end{tabular}                    \\ \hline 
\cite{s49}  & \begin{tabular}[c]{@{}l@{}} Secure mobile \\ edge  caching \end{tabular}                & \begin{tabular}[c]{@{}l@{}} Maximizing the \\  offloading traffic \end{tabular}                            & A3C                                                              & \begin{tabular}[c]{@{}l@{}} Network conditions \\  and signals \\ characteristics, i.e.,  \\ user requests and  \\ information \end{tabular} & \begin{tabular}[c]{@{}l@{}} Caching policy \\ (i.e., caching or not)\end{tabular}                                                                       & \begin{tabular}[c]{@{}l@{}} Computed based on the \\ the amount of traffic \\ at each edge node \end{tabular}                                            \\ \hline
\begin{tabular}[c]{@{}l@{}} \cite{s147}, \\ \cite{s148} \end{tabular} & Spoofing detection                                                                     & \begin{tabular}[c]{@{}l@{}}Create an optimal\\  authentication\\ model \end{tabular}                                              & \begin{tabular}[c]{@{}l@{}}Q-learning\\ and Dyna-Q\end{tabular}  & \begin{tabular}[c]{@{}l@{}} False alarm rate,\\ missing spoofing \\  rate,  and \\  detection rate \end{tabular}   & \begin{tabular}[c]{@{}l@{}}Authentication\\  discrete levels\end{tabular}                                                                       & \begin{tabular}[c]{@{}l@{}}Defined using an utility \\ function that is  \\ computed based on the \\ Bayesian  risk \end{tabular}                             \\ \hline
\cite{s46}       & \begin{tabular}[c]{@{}l@{}} Secure network \\ resources \\ allocation\end{tabular}      & \begin{tabular}[c]{@{}l@{}}Assign resources\\  to each user  \end{tabular}                                                            & \begin{tabular}[c]{@{}l@{}}Double\\   dueling\\ DQN\end{tabular} & \begin{tabular}[c]{@{}l@{}} Base stations' status,\\  caching contents, \\  and MEC servers' \\   conditions  \end{tabular}                             & \begin{tabular}[c]{@{}l@{}} Users to  base  \\ station association, \\ caching locations, \\ and selected MEC  \\ servers for  \\ computing \end{tabular} & \begin{tabular}[c]{@{}l@{}}Computed based on\\ the SNR of the wireless\\ access link, cache status, \\  and the computation \\ ability \end{tabular} \\ \hline
\cite{s81}       & Secure CPS                                                                             & \begin{tabular}[c]{@{}l@{}} Detecting  false\\  inputs\end{tabular}                                                                  & \begin{tabular}[c]{@{}l@{}} DDQN\\ and A3C\end{tabular}     & System output                                                                                                               & \begin{tabular}[c]{@{}l@{}}Select the next\\  input value\\ from a set of\\  constant inputs\\ signals \end{tabular}                              & \begin{tabular}[c]{@{}l@{}} Defined based on \\ the past dependent \\ life long property, \\ time and\\ output signal \end{tabular}        \\ \hline 
\end{tabular}
\end{table*}

\subsection{ Dynamic Treatment Regimes  \label{sec:Regimes}}

RL-based solutions have proved their efficiency in different aspects of  healthcare systems. In particular, different RL schemes have been applied for diverse dynamic treatment regimes, including chronic diseases, mental diseases, and highly infectious diseases \cite{h165, h167, h168, h170, h171, h172}.   
Indeed, dynamic treatment regimes map individualized treatment plans into  sequences of decision rules for each stage of clinical intervention, while considering current patients' state to support them with a recommended treatment. 
Such a sequential decision making process is perfectly fit with RL concept, since it can be easily modeled as an MDP, while finding the optimal treatment regime using several RL algorithms.  
Table \ref{tab:diseases} summarizes some of the e-health systems that exploited RL for addressing the challenges of different diseases. 
It is not the objective of this paper to provide an in-depth technical comparison on  different proposed e-health systems. However, we investigate the practical benefits of leveraging RL in such applications. In what follows, we group these e-health systems under two main categories: remote monitoring applications and hospital-based applications    

\subsubsection{Remote monitoring applications}	

Chronic diseases are defined broadly as conditions that last one year or more and require ongoing medical attention, which may limit patients' activities in daily living. Statistics show that chronic diseases constitute a considerable portion of death every year \cite{h80}. The widely prevailing chronic diseases include endocrine diseases (e.g., diabetes and hyperthyroidism), cardiovascular diseases (e.g., heart attacks and hypertension), various mental illnesses (e.g., depression and schizophrenia), cancer, obesity, and other oral health problems \cite{h81}.  
The main challenge of chronic diseases is that they require continuous monitoring for the patients' state, in addition to a sequence of medical intervention to avoid adverse effects of persistent treatment. Patients' conditions (such as treatment duration and dosage, as well as patients' response to certain medications) have to be continuously revised and updated to provide sufficient treatment.

 To relieve the pressure on healthcare facilities, it is important to widely rely on remote monitoring applications, which allow for moving large number of patients with mild symptoms into home care \cite{abdellatif2020edge}. In this context, RL schemes have been widely applied for supporting dynamic treatment regimes, inside and outside healthcare facilities, which helps in many chronic diseases and mental illnesses, as will be shown below. 

Diabetes is one of the main chronic diseases in the world that requires continuous-remote monitoring without limiting patients' activities.  According to \cite{108},  451 millions of people are living with diabetes in 2017. RL for patient specific glucose regulation in  artificial pancreas (AP) \cite{113} has attracted much interest recently.  
For instance, the authors in \cite{Diabetes2020} develop a DRL model for optimizing single-hormone (insulin) and dual-hormone (insulin and glucagon) delivery. In particular, the proposed model is developed using double Q-learning with dilated recurrent neural networks to provide effective closed-loop control of blood glucose levels for Type 1 diabetic patients. The presented results depict the efficiency of the proposed model in obtaining an effective glucose control strategy compared to standard basal-bolus therapy with low-glucose insulin suspension.  
In \cite{114}, RL is leveraged to control the blood glucose levels for Type 1 diabetic patients who undergo to an intensive insulin treatments. In particular, model-free Q-learning algorithm is used for insulin regulation rate and the generation of  the glucose-insulin dynamics. In \cite{130}, an  MDP model for patient state progression is presented. The reward for each treatment is computed based on the opinions of a doctor. 
To define a personalized treatment plan, a RL-based method (i.e., called V-learning), which proposes defining the optimal policy among a prespecified class of policies, is presented in \cite{132}. This methods aims to  estimate the diabetes treatment regimes while reducing the number of hypo and hyperglycemic episodes for patients with type 1 diabetes.  

Anemia is a common comorbidity in chronic renal failure, which has been also tackled using diverse RL models. 
In \cite{133}, the potential of RL in the management of renal anemia is studied. The proposed model aims to  control the patient's hemoglobin (HGB) levels using erythropoietin (EPO) values. Also, as the iron storage in the patient has an impact on the process of red blood cell creation, it is considered as a state component together with HGB. 
 
RL has been also applied for mental diseases, such as epilepsy\footnote{Epilepsy is one of the major severe neurological diseases that  affects around 1\% of the world population.} \cite{RL_Seizure2013}, depression, schizophrenia and various kinds of brain disorders.
For instance, a RL-based approach is leveraged in \cite{guez2010adaptive, guez2008adaptive} for dynamically learning an optimal neurostimulation strategy for the treatment of epilepsy.  
The proposed model extracts the  EEG signal as a 114-dimensional feature vector then applies the DRL approach for epilepsy detection. The presented results show that RL-based solution outperforms the related stimulation strategies in the literature and reduces the incidence of seizures by 25\%,  and total amount of electrical stimulation to the brain by a factor of 10. 
A computational model that captures the transition from inter-ictal to ictal activity is presented in \cite{h160}. The naive Q-learning method is used to optimize stimulation frequency for controlling seizures with minimum simulations.
In \cite{RL_Seizure2013},  the EEG recordings are classified into normal or preseizure baseline patterns using a K-nearest-neighbor (KNN) classifier. Then, a RL approach is used to adaptively update the normal and preseizure baseline patterns according to the feedback from the prediction result. This work is among the first to study the use of RL in seizure prediction.  

\begin{table*}[htp]
	\centering
		\caption{Summary of the relevant e-health systems.}
	  \label{tab:diseases}
		\begin{tabular}{|c|c|c|c|c|} 
			\hline 
		\textbf{Application} &\textbf{Medical Data} &\textbf{Disease} & \textbf{Description} & \textbf{RL Benefits}\\
		\hline
	Remote monitoring & insulin and  & Type 1 & DRL model is proposed for optimizing & The proposed method clearly   \\
	\cite{Diabetes2020}& glucagon & diabetes &basal insulin and glucagon & enhanced glycemic outcomes in  \\
		& levels &   & delivery & adult population  \\	
		\hline
		Remote monitoring&Glucose&Type 1 &The proposed method optimizes the dynamic & RL could improve patients' state \\	
		\cite{132}&level&diabetes&treatment regime for patients with &  by enabling frequent treatment  \\
		&   &   &type 1 diabetes & adjustments based on the evolving   \\
		&   &   & & health conditions of each patient \\
		\hline
		Remote monitoring&HGB&Anemia & It addresses the challenges of & The proposed approach generates \\
		\cite{133}&level&   &individual response variations to the treatment& adequate dosing strategies for \\
		&   &   &and changing patients' response over time & representative individuals from  \\
		&   &   & & different response groups \\		
		\hline 
		Remote monitoring&EEG&Epilepsy&An adaptive learning approach is proposed by&RL prove its efficiency in improving\\
		\cite{RL_Seizure2013}&   &&integrating RL with online monitoring to&prediction accuracy of the system \\
		&   &   & enhance personalized seizure prediction&   \\	
		\hline
		Remote monitoring &  EEG & Epilepsy &It obtains the optimal deep-brain & The proposed RL method could   \\
		\cite{guez2008adaptive}&&&stimulation strategy as a function of the & minimize the frequency and duration   \\
		&   &   &  acquired EEG signal & of seizures  \\	
		\hline
		Hospital-based & Dosage in &Cancer&A model-free TD method with Q-learning is & RL could obtain the optimal   \\	
		\cite{83}&cancer && applied for discovering individualized  & treatment strategies directly from  \\
		&chemotherapy & & treatment regimens & clinical data without  identifying  \\
		&   &   &   &  any accurate mathematical models  \\
		\hline
		Hospital-based&Drug dose &Cancer&Drug dose is fed as an input to the RL scheme,  & The obtained results showed that \\	
		\cite{85}&in cancer   && while defining the reward as a function of drug & RL yield better performance than     \\
		& chemotherapy  && dose and cell populations & conventional policy of pulsed   \\ 
		&   &   &   &  chemotherapy \\ 
		\hline
		Hospital-based & Drug dose&Cancer & Q-learning algorithm is proposed to  & This work states the power  \\	
		\cite{94}&in cancer && optimize dose calculation in cancer & of RL in optimizing complex      \\
		&radiotherapy&&  radiotherapy & biological problems such as   \\
		&   &   &   & radiotherapy in cancer treatment \\
		\hline
		Hospital-based& EHR&Sepsis& RL-based model is developed to & Mortality rate decreased in the \\	
	\cite{h193}&&& dynamically obtain the optimal treatments& patients that follow the treatment \\
		&   &   & for patients with sepsis in the ICU &  decisions obtained by the RL \\
		\hline
	Hospital-based&EHR& anesthesia & RL is utilized to design a & RL show comparable performance  \\	
	\cite{h213}&&& closed-loop anesthesia controller monitor& with the recent clinical  \\
		&   &   &and control the infusion of anesthetics& trials conducted \\
		\hline
		\end{tabular}
\end{table*}

\subsubsection{Hospital-based applications}

DRL techniques have been used for life-threatening diseases such as cancer\footnote{Cancer is one of the  most serious chronic diseases that cause death. Statistics show that 15.7\% of total deaths in the world is caused by  cancer.}.  
In \cite{83}, a model-free TD method with Q-learning is applied for agent dosage in cancer chemotherapy. A mathematical model for chemotherapy has been derived and two machine learning approaches (i.e., support vector regression (SVG) \cite{104} and extremely randomized trees (ERT) \cite{41}) have been used to fit the approximated Q-functions to the generated trial data. We show that reinforcement learning has tremendous potential in clinical research because it can select actions that improve outcomes by taking into account delayed effects even when the relationship between actions and outcomes is not fully known. 
In \cite{85}, a NAC (natural actor-critic) approach \cite{21} is proposed for optimizing the cancer drug scheduling policy. The main goal of the proposed approach is minimizing the tumor cell population and the drug dose while obtaining sufficient population levels of normal cells and immune cells. The obtained control policy by the proposed NAC approach suggests the drug should be injected continuously from the beginning until an appropriate time. This policy depicts better performance than traditional pulsed chemotherapy strategy. 

In \cite{94}, an agent-based simulation model along with a Q-learning algorithm are proposed to optimize dose calculation in cancer radiotherapy. Indeed, RL is used to optimize the main two factors of radiotherapy, i.e., radiation  dose  and  fractionation  scheme. In particular, each fraction is considered as a state variable, and the intensity of radiation is considered as an action variable, while presenting the reward as a function of the positive effect on tumor and negative effect on healthy tissue.    
The authors in \cite{87} present a Q-learning algorithm that aims to implement an optimal controller for cancer chemotherapy drug dosing.   
A DRL-based framework has been proposed in \cite{93} to automate the adaptive radiotherapy decision for non-small cell lung cancer patients. Three neural network components have been considered: (i) Generative Adversarial Network (GAN), which is used to generate sufficiently large synthetic data from historical small-sized real clinical data; (ii) Deep Neural Network (DNN), which is employed to learn how states would transit under different actions of dose fractions leveraging the synthesized data and the available real clinical data; and (iii) Deep Q-Network (DQN), which is responsible for mapping different states into possible dose strategies while optimizing future radiotherapy outcomes.  
In \cite{LymphNode}, a DRL-based solution is presented to detect abnormalities in medical images. In particular, the proposed scheme is used to optimize the lymph node bounding boxes, hence enhancing the lymph node segmentation performance, which is crucial for quantitatively accessing disease progression.   


RL is gaining also much interest for intensive care (IC) diseases \cite{h192, h193, h194, h200}. Early  detection and description of critical diseases significantly help doctors in the treatment, which can save many lives. Indeed, RL has been widely applied in the treatment of sepsis, regulation of sedation, and some other diseases that require intensive care unit (ICU) such as mechanical ventilation and heparin dosing. 
Sepsis is one of the main causes of death in hospitals and stands in the third place of mortality worldwide, however the optimal treatment strategy is still ambiguous \cite{h201, h197}. There is no universally agreed-upon treatment for sepsis due to the varying patients' responses to medical interventions.  
To tackle such complex decision-making problem, RL has been used in \cite{h193} to learn the optimal treatment policy by analyzing the massive amount of information extracted from different patients. This work depicts that the obtained treatment decisions by RL are on average more reliable than the decisions of human clinicians.  
In \cite{ h194, h195}, a DRL with continuous state-space model is used to deduce the optimal
treatment strategies from the available training samples that do not represent the optimal behavior. 
Indeed, the patient's physiological state is represented as a continuous vector (using physiological data from the ICU) in order to define the suitable actions with Deep-Q Learning. 
The application of RL for sepsis treatment has been also considered in \cite{h200}. However,  the authors there focus on the reward learning problem of RL. In particular, a Deep Inverse RL with Mini-Tree (DIRL-MT) model is presented to define the best reward function from a set of possible treatment strategies using real-world medical data. In this scheme, Mini-Tree model is used to learn the main components that affect the death rate during sepsis treatment, while deep inverse RL model being leveraged to obtain the complete reward function as a function of the weights of those components.  
In \cite{h198}, sepsis challenges have been tackled using mixture-of-experts framework to personalize sepsis treatment. Specifically, the proposed mixture model automatically swaps between neighbor-based (kernel) learning and DRL based on the current history of the patients. 
The obtained results in \cite{h198} depict the superior performance of the  proposed mixture model compared with applying the strategies of physicians, Kernel learning only and DRL only. 
In \cite{h203}, the authors leverage DDPG scheme to deal with the continuous state and action spaces of the sepsis environment, hence defining an effective treatment strategy for sepsis. 

It is defined  as a state of controlled, temporary loss of sensation. 
Critically ill patients, who are supported by mechanical ventilation, require adequate
sedation for several days to guarantee safe treatment in the ICU \cite{h235}. 
RL methods has been applied for the regulation of sedation in ICUs. The problem of anesthesia control using RL has been studied in several works such as \cite{h210, h213, h211}. 
A closed-loop anesthesia controller model is presented in \cite{h213}. This model is able to  regulate the bispectral index (BIS)  and mean arterial pressure (MAP) within a desired range. 
Specifically, a weighted combination of the error of the BIS and MAP signals is considered in the proposed RL algorithm. This reduces the computational complexity of the RL algorithm and consequently the controller processing time.   
In \cite{h211}, an Adaptive Neural Network Filter (ANNF) is proposed to improve RL model for closed-loop anesthesia control. In particular, the authors propose a method for smoothing the acquired BIS measurements while minimizing time delay. This leads to enhancing the patient state estimation, which allows for improving the performance of anesthesia controller. 
In \cite{h210}, an RL-based fuzzy controllers architecture is proposed for an automation of the clinical anesthesia. The authors presented a multi-variable anesthetic mathematical
model to compute the anesthetic state using two anesthetic drugs of Atracurium and Isoflurane.  


\begin{figure*}[t!]
\center{\includegraphics[width=3.4in]{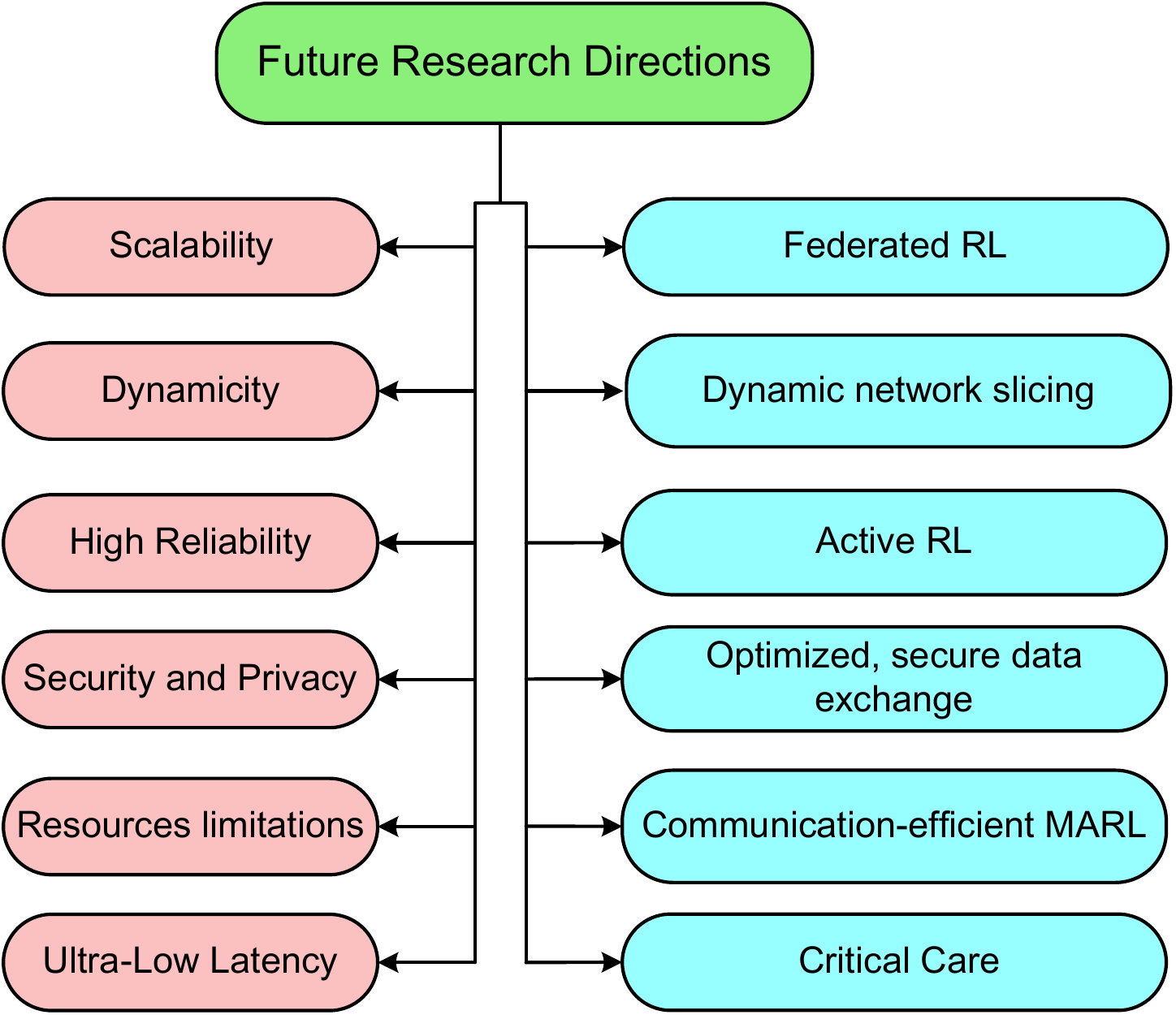}}
\caption{The proposed future research directions.  }
\label{fig:Future}
\end{figure*}

\section{Open Challenges and Future Research Directions \label{sec:Research} } 

In this section, we propose several open research directions along with their inherent challenges that can be tackled in future studies for implementing an efficient I-health system.  In particular, we investigate the opportunities of integrating the emerging RL models in new scenarios/services related to I-health system, which are  envisioned to emerge in future healthcare systems. 
Figure \ref{fig:Future} summarizes the main proposed future research directions. 

\subsection{Federated RL} 
Towards 6G networks evolution, edge intelligence is foreseen to be a crucial component in 6G network architecture. The concepts of AI for Edge Computing (EC) and EC for AI are emerging for the next generation networks. The former refers to implementing distributed computing applications (i.e., AI or deep learning techniques) at the MEC servers to enhance the performance of EC \cite{loven2019edgeai}, while the latter refers to leveraging EC capabilities to run sophisticated AI models. 
Despite the potentials of AI for EC, there are pivotal challenges that need to be address, including:
\begin{itemize}
  \item the complex structure of deep learning models and the need for large amount of data and resources for the training. Most of the existing systems that leverage deep/machine learning algorithms assume easy availability of the data. However, for example in E-health systems, the data is locally generated and stored across different devices/nodes in the network, e.g., PDAs, hospitals, private clinics, etc. Moreover, the locally generated data at each node cannot be gathered in a central node due to the privacy issues and rising demand for network bandwidth. 
  \item The computational capabilities of the mobile edge nodes are typically limited. Moreover, transferring large amount of data between the edges while considering the high-dynamic nature of wireless networks and restricted bandwidth is not an easy task \cite{abbas2017mobile}. 
  \item Traditional learning algorithms, that run independently at each node, cannot provide the required system level scalability and optimal performance. 
  \end{itemize}
  
To address these challenges, collaborative learning approaches (such as distributed training over multiple helpers and distributed inference among multiple participants) have been recently proposed. 
However, how to efficiently integrate collaborative learning approaches with MEC architecture in ultra-dense 5G network is a critical problem in the future development of 5G \cite{9205252}. 
Indeed, efficient utilization and collaboration between multiple edge nodes within 5G ultra-dense network should be considered in order to guarantee: 
\begin{itemize}
  \item efficient utilization of diverse resources over 5G heterogeneous network, such as computation, communication, and storage resources \cite{liu2020resource, 8767017};
  \item minimizing the overhead of decision-making and resource allocation processes; 
\item privacy and security protection for the acquired data.
\end{itemize}
The main question now is how to obtain the global optimal strategy for all collaborative edges? Unfortunately, due to the dynamic nature of the MEC systems and diverse characteristics of the collaborative edges, the generated optimization problems are always non-convex and NP-hard \cite{zhao2019computation}. 
Hence, it will be hard to rely on the traditional vanilla optimization methods for solving the joint MEC collaboration problems. 

\begin{figure}[t!]
\center{\includegraphics[width=3.4in]{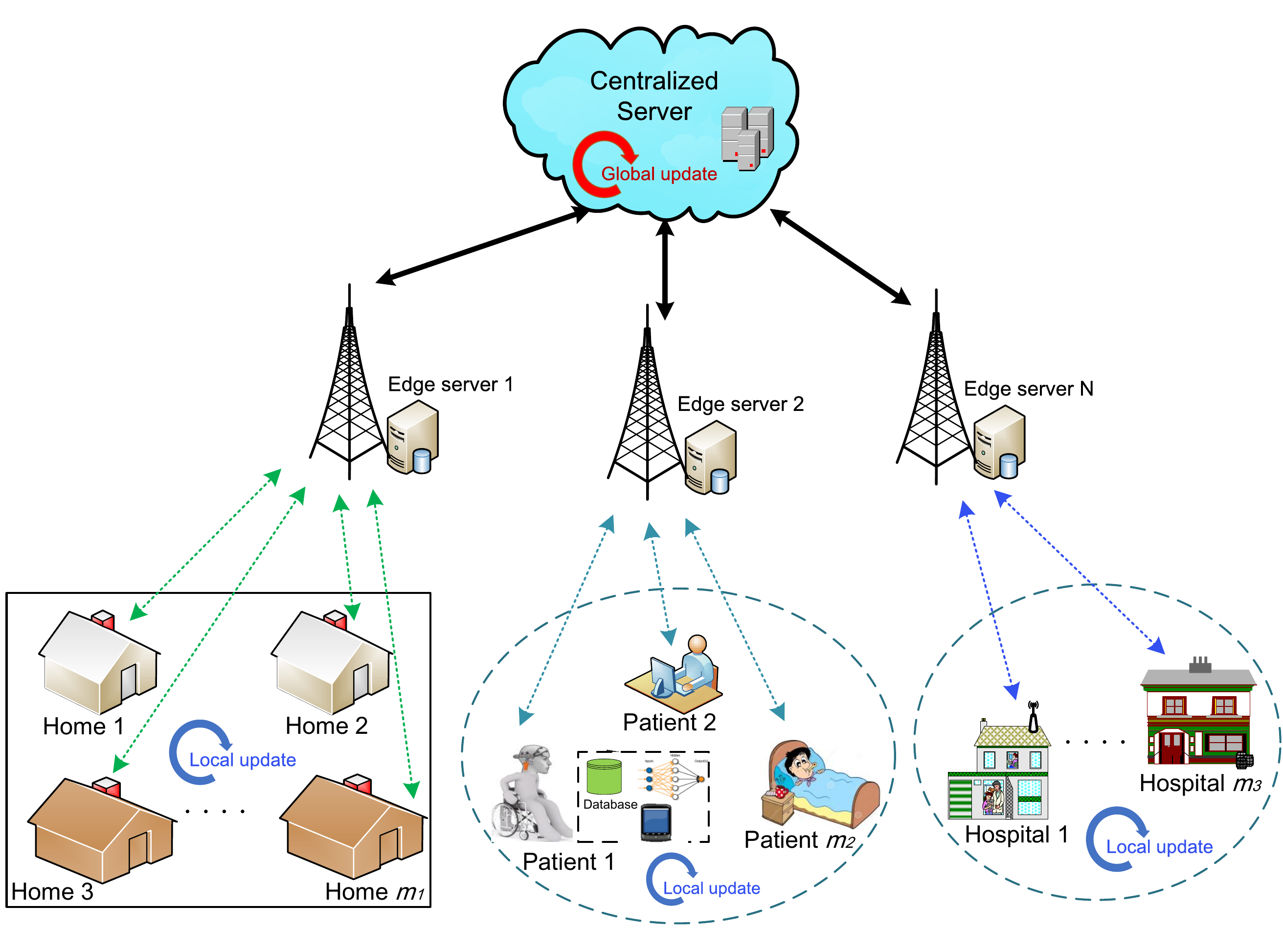}}
\caption{Hierarchical federated RL architecture for remote health systems.  }
\label{fig:FRL}
\end{figure}

Federated RL begins to attract more attention for optimizing the distributed/collaborative learning approaches, where an agent learns and obtains its optimal policy leveraging its local observations and cooperation with other agents for optimizing the same system targets \cite{zhu2021federated, 9247266, 9434397, 9447004, 9237167}. 
In Federated RL, a DRL agent build its own learning model by interacting with its environment. Then, it uploads the generated local model to a Central node that is responsible for aggregating the local models from different agents and generating a global model for all agents. This model is then broadcast to all DRL agents that exchange their previous local models with the updated global model and iteratively reconstruct their local models. 
The work in \cite{9349772} is the only study that considered federated RL in E-health system,  where a DDQN model is implemented at each edge node to define a stable and sequential clinical treatment policy, while extracting the knowledge from
electronic health records (EHRs) across all edge nodes by using a decentralized federated framework.  
Thus, we envision that the potentials of federated RL in healthcare systems is still worth further investigation. 
In particular, with the rapid interest in enhancing home care services, federated RL can be an efficient candidate for locally processing the patients' data and identify their states. Indeed,  leveraging federated RL solution within the large-scale healthcare systems is a key for detecting and managing urgent outbreaks, since it enables a swift and portable emergency detection through identifying and monitoring infected individuals at the edge, without the need of transferring patients' data to a centralized server (see Figure \ref{fig:FRL}).   

\subsection{Dynamic network slicing for I-health systems}

Strict latency and reliability  requirements needed for diverse healthcare applications cannot
be accomplished through the traditional telecommunication systems. The demanding need for higher data rates and extremely fast response time enable swift and accurate clinical decisions needed for patients' remote monitoring.   
Network slicing is a promising technology that can fulfil these strict requirements for healthcare application via dynamic slice allocation and isolation. 

Different from the related work in the literature, leveraging the intelligence at the edge for optimizing the  slice allocation decisions in ultra-dense heterogeneous networks is still an open research direction. Although few research attempts have been presented for dynamic network slicing using Q-learning method or RL \cite{9223714, 9131779}, we are still at the beginning level. Thus, a potential future direction can advance the state-of-the-art by:
\begin{itemize}
  \item leveraging RL algorithms with feature-assisted schemes to obtain swift and efficient slice allocation decisions in high-dynamic environments;
  \item  considering the impact of the resources' quality, users' behavior, and applications' characteristics on dynamic network slicing, hence avoiding inefficient utilization of network slices.  
\end{itemize}  
Moreover, building reliable prediction models to foresee users' demand and network dynamics, based on historical data and real-time gathered data, can significantly improve network slicing design through:  
\begin{enumerate}
    \item  stabilizing wireless connections,  
     \item optimizing radio resource management, and
  \item implementing proactive network slicing strategies. 
\end{enumerate} 

We argue that network dynamics are not the only pivotal factor for designing optimal, dynamic network slicing schemes. Predicting users' state (or behavior) can also be crucial for stabilizing wireless connections and fulfilling strict QoS requirements. Specifically, in E-health remote monitoring applications, the acquired medical data of the patients must be sent to the health cloud every 5-10 minutes, in normal conditions. On the contrary, in robotic telesurgery, the acquired physical vital signs must be reported with a latency less than 250 ms \cite{zhang2018towards}. By knowing such information in advance,  the network operator can always allocate the slice with low energy consumption in normal case, while allocating the slice with high data rates in case of emergency.  Thus, predicting users' behavior or state at the edge can significantly changing slice allocation decisions.  
%

\begin{figure}[t!]
\center{\includegraphics[width=3.4in]{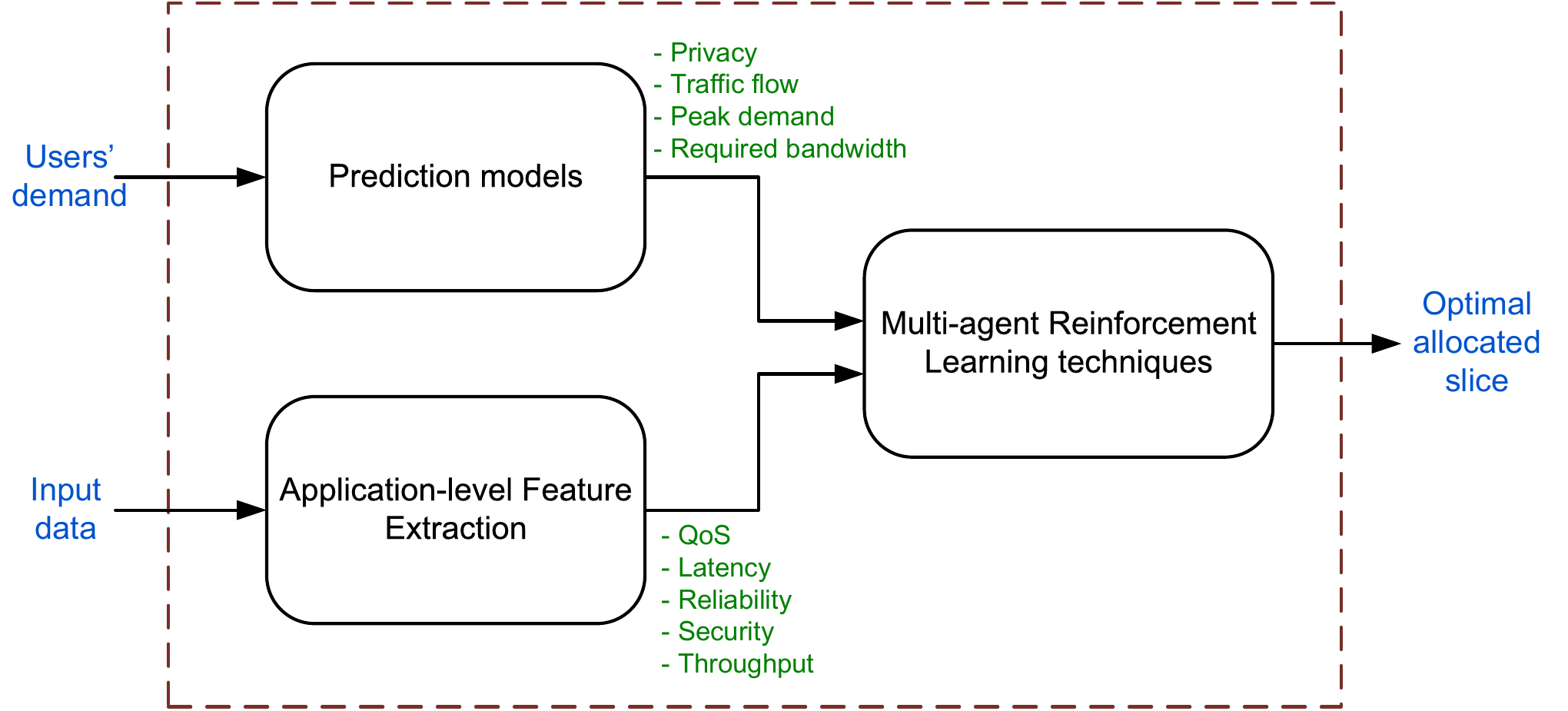}}
\caption{ An example for efficient network slicing solution over I-health system.   }
\label{fig:slicing}
\end{figure}

By accounting for the available slice characteristics, users' behavior and context, energy efficiency, and applications' requirements, efficient RL-based approaches can be developed to address the typical challenges of dynamic network slicing in I-health systems, using: 
\begin{itemize}
  \item historical information: the edge nodes can build efficient RL models for the surrounding-complex environment to accelerate the decision process of slice allocation based on their previous experiences;
\item  prediction models: by interacting with the surrounding environment, an edge node can utilize diverse performance indicators, such as the received power, received signal quality, and computational capabilities to predict the characteristics of different slices, hence connecting to the optimal slice; 
\item network dynamics: we can estimate at the edge the obtained reward before associating to a specific slice; 
\item applications' requirements: the edge node can associate to a customized slice based on the characteristics/requirements of the running applications. 
\end{itemize}
Accordingly, we envision that the effective solution should comprise two main steps (see Figure \ref{fig:slicing}): 
(i) estimating users' demand, resources availability and cost, and traffic flows using accurate RL models; (ii) developing MARL algorithms to be run at different edges for dynamic network slicing. 

\subsection{Active RL }

Home care applications have gained much interest recently. The goal of home care is to continuously monitor a large number of patients, at their homes, without limiting their daily activities. 
We envision that, patients, equipped with wearable devices, sensors, and PDAs, can learn, detect, and prevent complex situations occurring anytime anywhere, thanks to RL techniques. However, the challenge is that patients may be exposed to unexpected situations every day, such as emergency conditions, for which limited historical data is available. Moreover, most of RL techniques depend on a long training phase, where each agent has to interact and learn from the environment the optimal policy.     
However, we cannot rely on such an assumption in many real-time applications, such as critical healthcare applications. 
To address such a crucial problem, developing an effective solution that integrates RL with active leaning concept \cite{9162964},  will be needed. 
The main idea of AL is that an agent can actively interact with its neighbors to select over time the most informative data to be considered in the training phase in order to improve its learning
performance \cite{9380524}.     
Such  a solution that integrates the RL with the active leaning can comprise:  
\begin{itemize}
  \item leveraging Device-to-Device (D2D) communication to increase the amount of data acquired at a PDA \cite{7986262}; 
    \item investigating different types of data that can be exchanged between the PDAs, through D2D communication, and their impact on the obtained classification performance and communication load; 
  \item proposing efficient algorithms for data quality assessment and improvement, to enhance the quality of the collected data leveraging neighbors' information and data freshness; 
  \item developing efficient RL algorithms at the edge nodes (i.e., PDA). Specifically, upon the occurrence of unexpected conditions, connected PDAs do, not only interact with their environment, but also cooperate with their neighboring PDAs to optimize their rewards.  
\end{itemize}

\subsection{Optimized and secure data exchange} 

\begin{figure}[t!]
\center{\includegraphics[width=3.4in]{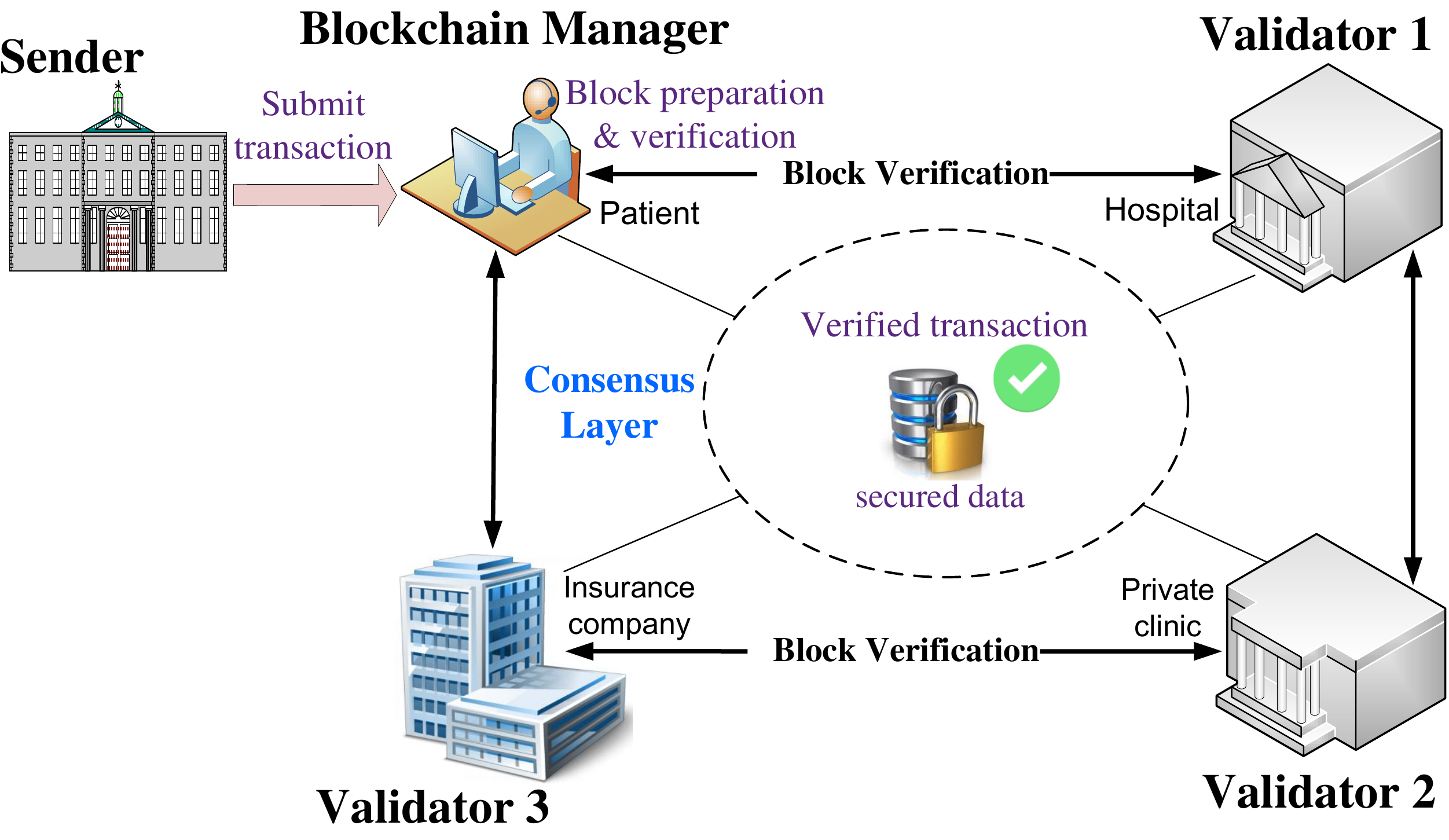}}
\caption{Blockchain architecture over I-health system.   }
\label{fig:Blockchain}
\end{figure}

Interactions and data exchange across distributed entities are essential to provide accurate control 
and  management,  and  improve  the  response  time  in  emergency  conditions.  However,  critical  challenges have emerged with data sharing and data access across distributed I-health systems, namely,
\begin{enumerate}
    \item data management in untrusted cloud servers, with risks for the users' privacy; 
     \item  fulfilling  diverse  security  and  privacy  requirements,  while  dealing  with  the  complexity  of  data  processing and transfer; 
      \item  remote accessibility of acquired data by different authorized entities. 
\end{enumerate} 
Thus, improving the accessibility and information sharing among diverse entities in I-heath system is mandatory to provide secure services (see Figure \ref{fig:Blockchain}). Indeed, we argue that a secure, trusted, and  decentralized  intelligent  platform  addressing  the  above  challenges  can  be  designed  and  realized  leveraging  edge  computing  and  blockchain  technologies.  Blockchain  is  a  decentralized  ledger  of  transactions that are shared among multiple entities while preserving the integrity and consistency of the data \cite{9091543}.  Being  decentralized, it well matches the potentiality of edge computing, which can effectively support data  storage and processing at different entities as well as their interconnection. 

Although blockchain promises to provide adjustable and distributed security protection for diverse healthcare applications \cite{8993839, Hyperledger_Fabric}, it poses also a number of challenges that need to be addressed. These include allowing a timely access to data for e-Health services, limiting latency, required storage space, computational power,  and  cost,  and  ensuring  data  privacy  protection.  Thus, developing an efficient RL-based solution to  address  these challenges  is needed. Indeed, RL-based solutions can adapt to different blockchain states and system's transitions in order to customize  and  optimize  the  blockchain  configuration's parameters  (e.g.,  number  of  verifiers,  block  size,  transaction size, block generation time out). Hence, such solutions would be able to establish the best  tradeoff among diverse conflicting objectives in I-health systems (e.g., latency, storage space, and cost) to support secure data sharing and storage services.   

\subsection{Communication-efficient MARL} 

The main wireless communication challenges, such as delay, noise, failure, and time-varying typologies, have been ignored in most of the studies that considered MARL.   
Although some of the recent studies have investigated the impacts of bandwidth and multiple-access techniques on the performance of emerging policies of MARL \cite{wang2020learning, mao2020learning}, it is still missing how to design effective MARL-based solutions under realistic wireless network constraints. 

Designing communication-efficient MARL solutions is still challenging. 
On one hand, algorithms that rely on training their models at centralized servers while running the execution tasks in a decentralized manner can reduce the communication overheads at the execution phase. Meanwhile, they will be able to obtain near-optimal joint policy at the centralized training phase. However, the adaptability of the obtained solution to the high-dynamics environments is not guaranteed. Moreover, re-training may be needed in some cases in order to adapt to the major changes in the environment.  
On the other hand, fully decentralized agents may be able to learn good policies, but this comes at the expense of communicating more often to react to their joint actions. 
Thus, it is important to design adaptable, yet communication efficient, MARL-based solutions,   which still needs further investigation.  

\subsection{Critical Care}  

Critical care is mainly related to seriously impacted patients who need swift and special medical treatments. Typically, intensive monitoring and fast reaction are needed for such patients, unlike the normal treatment for elders or chronic disease patients, who are usually less critical and need constant monitoring and medication for a long period of time \cite{vincent2013critical}.  
Remote surgery is one of the mission-critical healthcare applications that allows for providing a remote service from a set of consultants, who are in different places around the world, leveraging augment reality (AR) and virtual reality (VR). Since such type of applications directly deal with the patients' lives, they require communication services with ultra-reliability and ultra-low latency.  
Thus, medical data transmission and management for such critical care applications are very challenging research topics in healthcare. Much attention of future research topics should pave the way to provide high-quality healthcare-data transmission that satisfies diverse strict QoS requirements. RL algorithms can fulfil this gap by efficiently processing the acquired data in MEC servers, while minimizing the disturbance resulting from  the variation of network resources utilization from other applications.   
 


\section{Conclusion  \label{sec:Conclusion}  }

RL has gained an immense interest recently in several domains. Due to its efficiency in handling large-scale and highly-dynamic problems, different RL algorithms have been widely used in solving complex optimization problems. Thus, this paper presents a comprehensive review of the recent advances in RL models and their potential applications in I-health systems to fulfil diverse healthcare QoS requirements. 
Specifically, we first presented the major challenges in I-Health systems, while proposing our I-Health system architecture that aims at addressing these challenges. Then, we discussed the fundamentals and background of diverse RL, DRL, and MARL models, while highlighting the major benefits that can be obtained by incorporating such RL models within the proposed I-Health architecture. 
In addition,  we reviewed  the state-of-the-art applications of RL in three main areas: edge intelligence,
smart core network, and dynamic treatment regimes. 
In this context, many crucial challenges have been identified along with the RL-based solutions that have been proposed for, to mitigate these challenges in I-Health systems.
Finally, we presented our vision for the open challenges and problems that need to be tackled in the future, along with some research directions for innovation. 

\section*{Acknowledgment}
This work was made possible by NPRP grant \# NPRP12S-0305-190231 from the Qatar National Research Fund (a member of Qatar Foundation). The findings achieved herein are solely the responsibility of the authors.  




\balance
\bibliographystyle{IEEEtran}
\bibliography{ref2}

\end{document}